\pdfoutput=1

\documentclass[11pt]{article}

\usepackage[final]{acl}

\usepackage{times}
\usepackage{latexsym}

\usepackage[T1]{fontenc}

\usepackage[utf8]{inputenc}

\usepackage{microtype}

\usepackage{inconsolata}

\usepackage{graphicx}

\usepackage{subcaption}
\usepackage{mwe}

\usepackage{multirow}
\usepackage{microtype}
\usepackage{booktabs}

\usepackage{amsfonts}

\usepackage{float} 
\usepackage{amsmath}
\usepackage[colorinlistoftodos]{todonotes}
\usepackage{rotating}

\title{The Lookahead Limitation:\\ Why Multi-Operand Addition is Hard for LLMs}

\newcommand{\affilsup}[1]{\rlap{\textsuperscript{\normalfont#1}}}

\author{
    Tanja Baeumel\affilsup{1,2}
    \qquad
    Josef van Genabith\affilsup{1, 3}
    \qquad
    Simon Ostermann\affilsup{1,2}
    \\
    $^1$German Research Center for Artificial Intelligence (DFKI) \\
    $^2$Centre for European Research in Trusted AI (CERTAIN) \\
    $^3$Department of Language Science and Technology, Saarland University \\
    Saarland Informatics Campus, Saarbrücken, Germany\\
    \texttt{\{firstname.lastname\}@dfki.de} \\
}

\begin{document}
\maketitle
\begin{abstract}
Autoregressive large language models (LLMs) exhibit impressive performance across various tasks but struggle with simple arithmetic, such as additions of two or more operands. We show that this struggle arises from LLMs’ use of a \textbf{simple one-digit lookahead heuristic}, which works fairly well (but not perfect) for two-operand addition but fails in multi-operand cases, where the carry-over logic is more complex. Our probing experiments and digit-wise accuracy evaluation show that LLMs fail precisely where a one-digit lookahead is insufficient to account for cascading carries. We analyze the impact of tokenization strategies on arithmetic performance and show that all investigated models, regardless of tokenization, are inherently limited in the addition of multiple operands due to their reliance on a one-digit lookahead heuristic. Our findings reveal fundamental limitations that prevent LLMs from generalizing to more complex numerical reasoning.
\end{abstract}
\section{Introduction}

Large language models (LLMs) demonstrate remarkable performance across a wide range of tasks \cite{bai2023qwen,gemmateam2024gemmaopenmodelsbased,guo2025deepseek}, yet consistently struggle with simple arithmetic tasks, such as the addition of multiple or large numbers \cite{mcleish2024transformers, shen2023positionaldescriptionmatterstransformers, zhou2023algorithmstransformerslearnstudy, zhou2024transformersachievelengthgeneralization}.

\begin{figure}[t]
    \centering
    \includegraphics[width=0.3\textwidth]{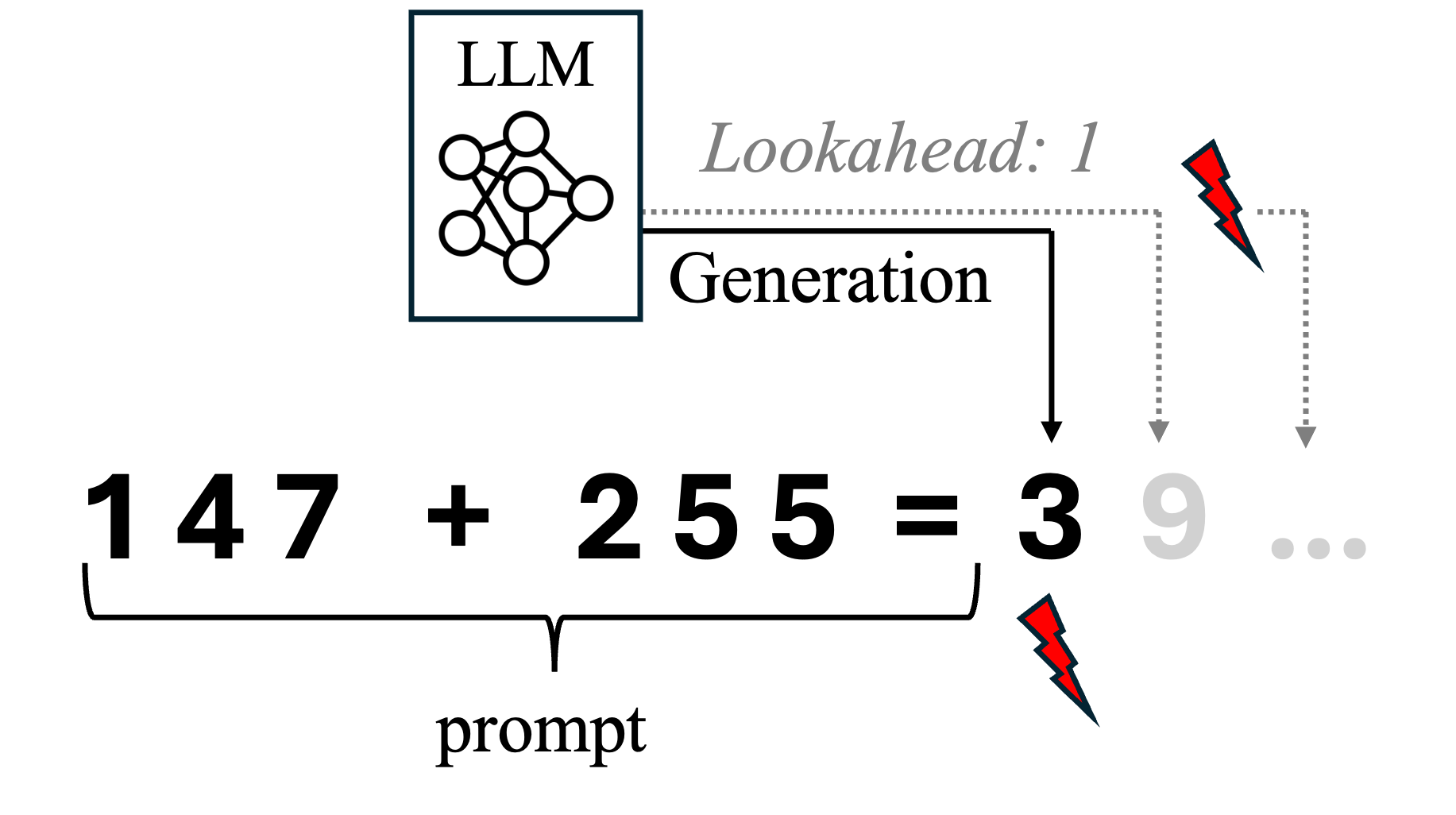}
    \caption{An addition of two three-digit operands.
    LLMs rely on a one-digit lookahead when performing addition. If a relevant carry emerges at a later stage in prediction, they fail to account for it, leading to errors in earlier generated result digits.}
    \label{fig:figure1}
    \vspace{-0.5cm}
\end{figure}

Figure \ref{fig:figure1} shows an example of an addition with 2 operands, $147$ and $255$, each with three digits ($0$ to $9$). The \textit{length} of an operand is the 
number of digits it contains. Figure \ref{fig:figure1} provides an example where the LLM 
fails (even in a two-operand case) to provide a correct output due to its insensitivity to a carry emerging from later computations.

The difficulty LLMs face in such tasks stems from the mismatch between the left-to-right nature of autoregressive language modeling and the right-to-left structure of standard arithmetic algorithms. Conventional addition methods process numbers digit by digit from right to left, propagating carries, while LLMs generate numbers sequentially from left to right without explicit intermediate calculations. This raises the question: What strategy do LLMs use to handle this misalignment in addition?

In this work, we show that in fact LLMs rely on a simple heuristic that enables high (though not perfect) accuracy in adding two operands
(e.g., $147 + 291 = 438$, henceforth \textit{two-operand addition}). This heuristic attempts to bridge the gap between the left-to-right generation
and the resulting need to 'look ahead' to account for propagating carries from less significant digits. Rather than performing an exhaustive lookahead to fully anticipate carry propagation, \textbf{LLMs rely on a simple heuristic that involves a lookahead of only a single digit to anticipate the value of carries in addition}. 
We show that while this strategy works fairy well
for two-operand addition, due to relevant digit combinatorics, it deteriorates substantially with multiple operands
(e.g., in four-operand addition such as $147 + 245+ 312 + 104 = 808$, henceforth generalized as \textit{multi-operand addition} for any number of operands $>2$), where anticipating carries becomes less predictable. The reliance on the heuristic explains the lack of robustness in LLMs' arithmetic performance. 

Figure \ref{fig:figure1} illustrates this shortcoming of the heuristic: A one-digit lookahead anticipates no carry (because for the sum of the second, i.e. middle, digits in the operands $4+5=9$), leading to the inaccurate prediction of the first result digit as $3$, unable to accurately anticipate the cascading carry originating from the unit position. 

To gather evidence that the heuristic accurately describes the strategy used by LLMs to solve addition from left to right, we present results from three state-of-the-art LLMs with different tokenization strategies (single digit and multiple digit) for numerical outputs. 
By evaluating prediction accuracy on carefully curated datasets and employing probing techniques, 
we provide multiple lines of evidence that LLMs struggle specifically with addition tasks where a one-digit lookahead is insufficient to account for cascading carries.
For instance, in two-operand addition, we show that this issue occurs when the sum of the digits at the lookahead position is $9$, leading to failure in correctly predicting the numerical value at the current position.
For example, in $147 + 255 =$, no carry is predicted for the middle digits, even though a cascading carry from the $10^0$ position affects the sum of the $10^1$ digits, and thus the $10^2$ position.

Our findings show that all investigated LLMs are inherently limited in their performance on multi-operand addition tasks due to this heuristic, regardless of their tokenization strategy.

Our contributions are as follows: 
\begin{itemize} 
\item \textbf{Evaluation of Addition Capabilities}: We show that LLMs fail on multi-operand addition (Section \ref{sec:acc_data}) and then systematically evaluate the capabilities of LLMs on two-operand addition tasks via probing (Section \ref{sec:probing}).
\item \textbf{Heuristic Discovery}: Inspired by results of the evaluation, we
formalize left-to-right addition in LLMs for multi-operand addition with a simple heuristic that uses a shallow lookahead of one to attempt left-to-right addition (\textbf{H1}, Section \ref{subsec:digit10}).
\item \textbf{Empirical Validation}: We demonstrate that \textbf{H1} is fragile in multi-operand addition and explain the performance decline as a function of the increasing number of operands in large comprehensive addition experiments. We find that model performance aligns \textit{precisely} with the predicted limitations of \textbf{H1} (Sections \ref{sec:h1_2op} and \ref{sec:multi_fail}). We find that \textbf{H1} holds independently of tokenization strategies (Section \ref{sec:llama}).
\end{itemize}

\section{LLMs Struggle with Multi-Operand Addition}
\label{sec:acc_data}

In this section, we define the data and models used in this work and demonstrate that LLMs fail on multi-operand additions by looking at prediction accuracy.

\subsection{Models and Data}
\label{sec:models_data}
\paragraph{Models.}
We compare Mistral-7B \cite{jiang_mistral_2023}, Gemma-7B \cite{gemmateam2024gemmaopenmodelsbased} and Meta-Llama-3-8B \cite{grattafiori2024llama3herdmodels, llama3modelcard} as they employ different tokenization strategies for numerical outputs: 
While Mistral and Gemma exclusively employ a single-digit tokenization strategy for their numeric input and generated output (e.g., input = ['1', '4', '7', '+', '2', '5', '5', '='], output = ['4', '0', '2']), Llama-3 employs a multi-digit numeric tokenization strategy (e.g., input = [' 147', ' +', ' 255', ' ='], output = [' 402']), typically favoring numeric tokens of length 3. 
\paragraph{Data.}
For all experiments in this paper, we compile a range of datasets containing simple arithmetic task prompts of the form \textit{147 + 255 = }. We create a dataset for each addition task ranging from 2-operand to 11-operand addition, where each operand is a triple-digit number between 100 and 899. Each of the 10 datasets contains 5,000 unique arithmetic problems, both in a zero-shot and one-shot setting. In the zero-shot setting, an example for a 2-operand addition prompt is ``147 + 255 = ''. An example for a 4-operand addition prompt is ``251 + 613 + 392 + 137 = ''.  Our one-shot prompt template follows the scheme \textit{q1 r1; q2 }, e.g.~``359 + 276 = 635; 147 + 255 = '', where \textit{q1} is a sample query from the same dataset and \textit{r1} is the correct result of the addition task in \textit{q1}. \textit{q2} is the query containing the addition task to be solved.

In the remainder of the paper, we use $s_n$ (with $n\geq 0$) to denote the result digit generated at digit position \(10^n\).
For example, in ``147 + 255 ='', with expected output 402, $s_2 = 4$, $s_1 = 0$, and $s_0 = 2$. 

\subsection{LLM Accuracy on Addition Tasks}
\begin{figure}[t]
    \centering
    \includegraphics[width=0.42\textwidth]{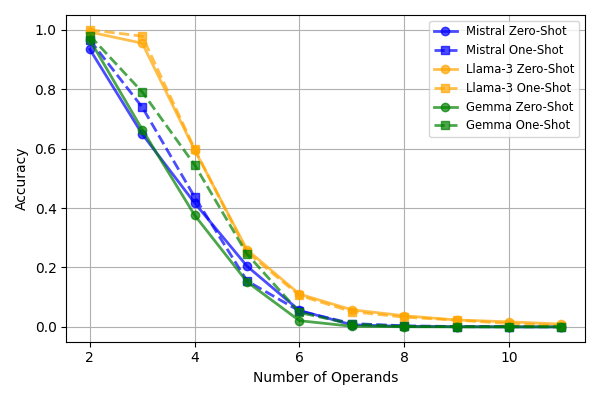}
    \caption{Accuracy of Mistral, Gemma and Llama-3 on multi-operand addition of triple-digit numbers, in a zero- and one-shot setting.}
    \label{fig:multi_op_accuracy_overall}
\end{figure}

Figure \ref{fig:multi_op_accuracy_overall} illustrates the significant decline in performance of Mistral-7B \cite{jiang_mistral_2023}, Gemma-7B \cite{gemmateam2024gemmaopenmodelsbased} and Meta-Llama-3-8B \cite{llama3modelcard} in multi-operand addition as the number of operands increases. This drastic decrease highlights the inability of these models to generalize effectively to addition tasks involving a higher number of operands, despite their strong overall capabilities. 

\section{Probing LLMs on Digits in Two-Operand Addition Tasks}
\label{sec:probing}

Solving arithmetic tasks presents a fundamental challenge for LLMs, as they generate text from left to right, while addition requires a right-to-left process due to carry propagation from the least significant to the most significant digit.
For instance, predicting the first result digit $s_2 = 4$ in ``147 + 255 = '' requires the model to anticipate that a carry originating from $s_0$ cascades through $s_1$ to $s_2$. Robust left-to-right addition thus requires a lookahead spanning all result digits, raising the question: Do LLMs internally represent future result digits when predicting $s_2$ - and if so, how far can they ``look into the future''?

To answer this question, we probe whether models accurately encode future result digits $s_1$ or $s_0$ while generating $s_2$. Building on \citet{levy2024language}, who show that, irrespective of a model's numeric tokenization strategy, LLMs internally represent numbers digit-by-digit in base 10, we analyze digit-wise probing accuracy on the two-operand addition dataset described in Section \ref{sec:models_data}.

\subsection{Methodology and Experiments} 
\paragraph{Data.}
We split the two-operand addition dataset (see Section \ref{sec:models_data}) into train (n=4500) and test (n=500) for the probing experiments. The two-operand addition dataset is designed such that correct results for the addition tasks are triple-digit numbers between 200 and 999. We use the zero-shot prompt setting for the probing experiment.

\paragraph{Probing Setup.}
Our goal is to determine which result digits are available at the prediction step of $s_2$. We thus train probes to predict the result digits $s_2$, $s_1$, and $s_0$ from hidden states of the model during the prediction step of $s_2$. 

Specifically, we train one-layer linear probes to predict individual digit values of the results from the hidden state of the last token at each model layer.  Probes are trained on the train split of the two-operand addition dataset and evaluated on the test split. 
We train separate probes to predict individual result digits $s_2$, $s_1$, and $s_0$, for all models at all layers.\footnote{We choose a low temperature of 0.1 during model inference to ensure deterministic and consistent outputs, reducing randomness in token generation and improving the reliability of numerical calculations.}

\begin{figure}[t]
    \centering
    \includegraphics[width=0.5\textwidth]{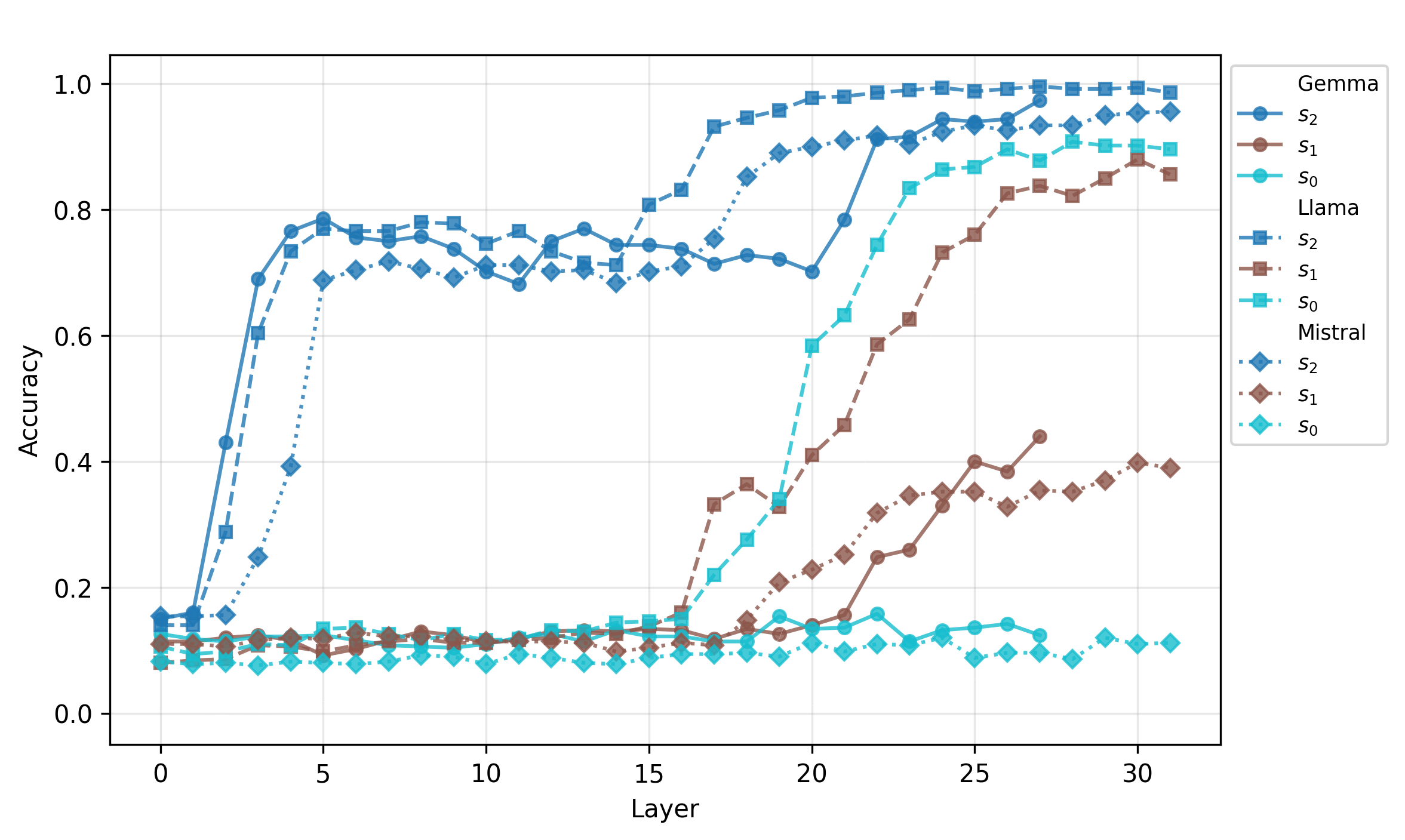}
    \caption{Probing accuracy of individual result digits as predicted by the hidden states of Mistral, Gemma and Llama-3. For two-operand, zero-shot addition prompts.}
    \label{fig:probing_multi_op_accuracy_overall}
\end{figure}

\subsection{Results}
\label{subsec:results}
The probing accuracy of individual result digits is shown in Figure \ref{fig:probing_multi_op_accuracy_overall}. Gemma and Mistral with their digit-wise tokenization internally represent only $s_2$ with high accuracy. In contrast, there is a high probing accuracy across \textit{all} result digits in Llama-3. This is due to the fact that Llama-3 tokenizes numbers into 3-digit numeric tokens: It is forced by its tokenization to generate all result digits ($s_2$, $s_1$, and $s_0$) in one step as a single token.

The single-digit tokenization models Mistral and Gemma exhibit a low probing accuracy on $s_0$ ($< 0.24$) in all layers.
Recall that $s_0$ is probed from the models' hidden states while they autoregressively generate $s_2$. 
We interpret the lack of internal representation of $s_0$ as evidence that these models disregard the potential influence of $s_0$ (including any cascading carry) when generating $s_2$.

In line with this, Gemma and Mistral show notably higher probing accuracy on $s_1$ compared to $s_0$, when probing from the models' hidden states as they generate $s_2$. 
We thus conjecture that the single-digit-token models seem to recognize the potential influence of the carry resulting from the sum of the \(10^1\) operand digits. Simply put, generating the digit at $10^2$ might employ a lookahead of one digit to the \(10^1\) intermediate result. 
Based on this observation, we formulate a hypothesis for a heuristic used by LLMs:
\begin{center}
    \textbf{H1: \indent LLMs employ a look ahead of one digit to generate the current digit of an addition task.}
\end{center}

\textbf{H1} would explain why LLMs cannot effectively represent each necessary digit of the result during generation, making it difficult to anticipate later carry values correctly. We first formalize \textbf{H1}, which explains the patterns observed in Figure \ref{fig:probing_multi_op_accuracy_overall}, in the next Section, and then verify the fit of \textbf{H1} with empirical addition outcomes generated by the models in Sections \ref{sec:h1_2op}, \ref{sec:multi_fail}, and \ref{sec:llama}.

\section{The Carry Heuristic of LLMs}
\label{subsec:digit10}

Since LLMs generate numbers from left to right, they must anticipate whether a carry from later digits (with lower bases further on in the result) will impact the current digit they are generating. In this section, we evaluate the maximum accuracy LLMs can achieve in addition tasks, assuming they rely on \textbf{H1}, given the limited lookahead of one digit.

\subsection{Formalization of Left-to-Right Addition in Base 10}
We first formalize a recursive algorithm for solving addition of $k$ operands-where each operand is a base 10 integer- in a left-to-right manner. 

\noindent \textbf{We define:}
\begin{itemize}
    \setlength{\itemsep}{0.2pt}  
    \setlength{\parskip}{0.2pt} 
    \item \( k \): Number of operands.
    \item \( n_1, n_2, \dots, n_k \): Operands, each represented as digit sequences in base \( 10 \), with \(\quad 0 \leq i < d\), where $d$ is the number of digits in the operands: \( n_j = [n_{j, d-1}, \dots, n_{j, 0}], \quad n_{j, i} \in \{0, \dots, 9\}\)
    \item \(S\): The result of the addition. \( S = [s_d, s_{d-1}, \dots s_0] \), where 
    \(s_d = c_d\), i.e., the final carry.
\end{itemize}

\noindent We recursively define the calculation of individual result digits:
\begin{itemize}
    \item \textbf{Total Sum at Digit Position \( i \):}
    \[t_{i} =\sum_{j=1}^k n_{j, i}\]
    \[ T_i = t_i + c_i\]
    where \( t_i \) is the digit sum at the current position, \( c_i \) the carry from the previous digit position, and $k$ the number of operands. Base case: \(c_0 = 0\), no carry at the least significant digit.
    \item \textbf{Result Digit at Position \( i \):}
    \[
    s_i = T_i \mod 10
    \]
    \item \textbf{Carry to the Next Digit Position:}
    \[
    c_{i+1} = \left\lfloor \frac{T_i}{10} \right\rfloor
    \]
\end{itemize}

A worked example is provided in Appendix \ref{appendix:A}.

\subsection{A Naive Heuristic for Solving Addition Left-to-Right}
Due to the recursive nature of left-to-right addition, a lookahead of \(i-1\) digits is needed to determine any result digit $s_i$. 
There is however a simple, non-recursive heuristic for the estimation of $s_i$ with only a one-digit lookahead, to the digit sum of the next position, i.e. only considering $t_{i-1}$. 

We define $c_{min}$ and $c_{max}$ to be the minimal and maximal possible value for a carry, where trivially for all cases, $c_{min}=0$, and 
\[c_{max}(k) = \left\lfloor \frac{\sum_{j=1}^k 9}{10} \right\rfloor\] 
in base $10$ and for $k$ operands. 
We then define the carry heuristic $c_{i}^{h}$ as follows: 
\[c_{i}^{h} \in \{ \left\lfloor \frac{t_{i-1} + c_{min}}{10} \right\rfloor, \left\lfloor \frac{t_{i-1} + c_{max}}{10} \right\rfloor \} \]  
Where $c_{i}^{h}$ is chosen uniformly at random. 
We then accordingly define the predicted total sum at digit position i
\[T_i^h = t_i + c_i^h\] 

and the predicted result digit

\[s_i^h = T_i^h \mod 10\]

\paragraph{Examples.}
We show two examples of two-operand addition, one in which \textbf{H1} is successful, and one in which it fails.
For $k=2$, i.e., in two-operand addition:
\[c_{max}(2) = \left\lfloor \frac{\sum_{j=1}^2 9}{10} \right\rfloor = 1\] 

\begin{figure}[t]
    \centering
    \includegraphics[width=0.372\textwidth]{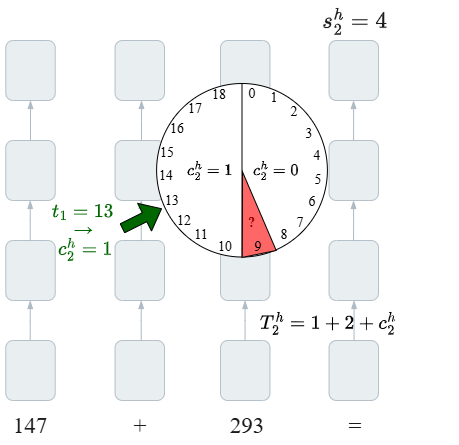} 
    \caption{Two-operand addition in which \textbf{H1} is successful.} 
    \label{fig:carry_2_op_success}
    \vspace{-0.2cm}
\end{figure}

\paragraph{147 + 293.} See Figure \ref{fig:carry_2_op_success}. We need $T_2^h$ and thus $c_2^h$ to generate the first result digit $s_2^h$. 
\[c_{2}^{h} \in \{\left\lfloor \frac{4 + 9 + c_{min}}{10} \right\rfloor, \left\lfloor \frac{4 + 9 + c_{max}}{10} \right\rfloor \}\]
\[=\{ \left\lfloor \frac{13}{10} \right\rfloor, \left\lfloor \frac{14}{10} \right\rfloor \} = \{1, 1\}\]
therefore $c_{2}^{h} = 1$, $T_2^h = 4$, and $s_2^h = 4$. \textbf{H1} succeeds in predicting the first digit $s_2$ for \textbf{147 + 293}. 

\begin{figure}[t]
    \centering
    \includegraphics[width=0.37\textwidth]{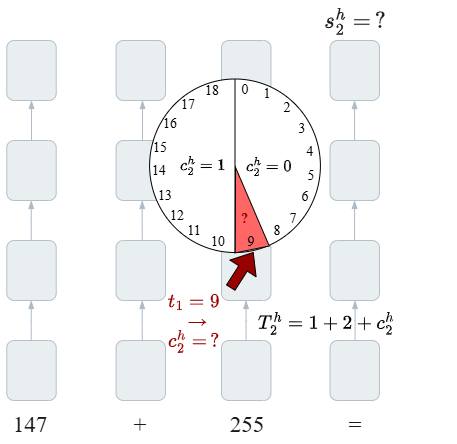} 
    \caption{Two-operand addition in which \textbf{H1} fails.} 
    \label{fig:carry_2_op_fail}
    \vspace{-0.2cm}
\end{figure}

\paragraph{147 + 255.} See Figure \ref{fig:carry_2_op_fail}. \\
\[c_{2}^{h} \in \{ \left\lfloor \frac{4 + 5 + c_{min}}{10} \right\rfloor, \left\lfloor \frac{4 + 5 + c_{max}}{10} \right\rfloor \}\]
\[= \{ \left\lfloor \frac{9}{10} \right\rfloor, \left\lfloor \frac{10}{10} \right\rfloor \} = \{0, 1\}\]
therefore $c_{2}^{h}$ is chosen uniformly at random between $0$ and $1$.
The heuristic fails in predicting the first digit $s_2$ for \textbf{147 + 255} with a 50\% chance. 
\section{H1 Predicts Difficulties of LLMs in Two-Operand Addition}
\label{sec:h1_2op}
In this section we show that single-digit token LLMs struggle exactly in those cases in which the heuristic \textbf{H1} is insufficient. 

\subsection{Predicted Accuracy} For two-operand addition, there are 19 possible values for each $t_i$ (ranging from 0 to 18, because this is the range of sums between two digits). In 18 out of these 19 cases, \textbf{H1} reliably determines the correct carry value. Only if $t_i = 9$, \textbf{H1} must randomly choose between two possible carry values, thus failing with a 50\% chance. This results in an overall predicted accuracy of
\[\frac{18\times1.0 + 1 \times 0.5}{19} = 0.974\]
for the first result digit $s_2$ in two-operand addition: \textbf{H1} achieves 97.4\% accuracy in correctly predicting the first result digit $s_2$. This corresponds almost exactly to Gemma's and Mistral's accuracies for generating $s_2$ during zero-shot and one-shot inference (Gemma: 0-shot: $97.12\%$, 1-shot: $98.04\%$; Mistral: 0-shot: $94.60\%$, 1-shot: $97.46\%$). Table \ref{tab:gen_accuracy_all} in Appendix \ref{sec:appendix_F} provides all generation accuracies for the data described in Section \ref{sec:models_data}.

\subsection{Finegrained Analysis} 
We further investigate whether it is true that especially cases with $t_i=9$ are challenging for LLMs. 

\paragraph{Data.} To this end, we evaluate prediction accuracy across five distinct newly introduced datasets, each containing 100 queries with distinct carry scenarios. The datasets follow the zero-shot template described in Section \ref{sec:models_data} and are designed to exhaustively capture all cases of carries affecting $s_2$ in two-operand addition of triple-digit numbers. 
\begin{itemize}
    \setlength{\itemsep}{0.1pt}
    \setlength{\parskip}{0pt}
    \setlength{\parsep}{0.5pt}
    \item \textbf{Dataset 1 (DS1): No carry.} The addition does not produce any carry (e.g., \(231 + 124 = 355\)).\footnote{We employ the additional constraint that the sum of the \(10^1\) operand digits $\neq 9$, i.e., ($s_1 \neq 9$)}. 
    \item \textbf{Dataset 2 (DS2): Carry in position \(10^0\), no cascading.} A carry is generated in the \(10^0\) ($s_0$) digit but does not cascade to the \(10^2\) ($s_2$) digit (e.g., \(236 + 125 = 361\)). 
    \item \textbf{Dataset 3 (DS3): Cascading carry from \(10^0\) to \(10^2\).} A carry originates in the \(10^0\) ($s_0$) digit and cascades to the \(10^2\) ($s_2$) digit (e.g., \(246 + 155 = 401\)).
    \item \textbf{Dataset 4 (DS4): Direct carry in position \(10^1\).} A carry is generated in the \(10^1\) ($s_1$) digit and directly affects the \(10^2\) ($s_2$) digit (e.g., \(252 + 163 = 415\)).
    \item \textbf{Dataset 5 (DS5): No carry, but position \(10^1\) digits sum to 9.} There is no carry in any digit, but the sum of the \(10^1\) operand digits is 9, i.e., ($s_1 = 9$) (e.g., \(256 + 142 = 398\)).
\end{itemize}
DS1 to DS5 can be neatly categorized according to whether the heuristic can accurately predict $s_2$: 

\begin{itemize}
    \item DS1 and 2: $t_1 = \sum_{j=1}^2 n_{j,1} < 9 \rightarrow c_{2}^{h} = 0$
    \item DS4: $t_1 = \sum_{j=1}^2 n_{j,i} > 9 \rightarrow c_{2}^{h} = 1$ 
    \item DS3 and 5: $t_1 = \sum_{j=1}^2 n_{j,1} = 9 \rightarrow c_{2}^{h} = ?$
\end{itemize}

\begin{figure}[t]
    \centering
    \begin{subfigure}[b]{0.33\textwidth} 
        \centering
        \includegraphics[width=\textwidth]{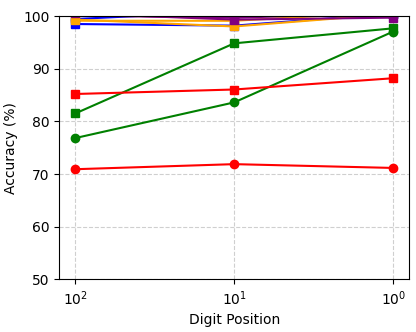} 
    \end{subfigure} 
    \begin{subfigure}[b]{0.09\textwidth}
        \centering
        \includegraphics[width=\textwidth]{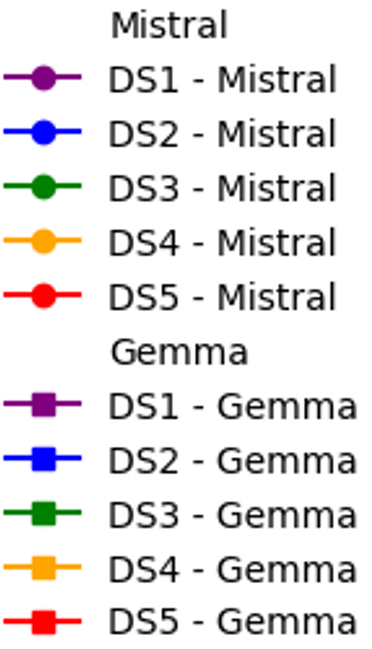} 
        \vspace{0.8cm}
    \end{subfigure}
    \caption{Per-digit generation accuracy of Mistral and Gemma on datasets DS1-DS5. Each dataset represents a different carry scenario.}
    \label{fig:scenario_mistral_gemma}
\end{figure}

\paragraph{Results.} Figure \ref{fig:scenario_mistral_gemma} shows that  LLMs struggle with DS3 and DS5, which are precisely the cases where \textbf{H1} predicts issues. As \textbf{H1} suggests, predicting the first result digit $s_2$ at position $10^2$ is particularly error-prone in these scenarios. 
The difficult datasets are the ones where a lookahead of one digit position does not suffice to determine the value of the carry needed to generate $s_2$. Simply put:
Overall, addition results tend do be predicted correctly by LLMs, if and only if a lookahead of one digit is sufficient to determine the value of the carry bit affecting $s_2$. Prediction is often incorrect if a lookahead of two or more digits is needed to determine the value of the carry bit affecting $s_2$.

In cases where a lookahead of one digit is enough to accurately determine the value of $s_2$ (DS1, DS2, DS4), the models succeed.
However, when a lookahead of one digit is insufficient to determine the value of $s_2$ (DS3 and DS5), the model struggles with predicting $s_2$ correctly.
Table \ref{tab:accuracy_2-op} in Appendix \ref{appendix:B} provides the generation accuracy of $s_2$ for Gemma and Mistral, in addition to the plot. 
Additionally, Appendix \ref{sec:appendix_G} presents probing experiments that yield the same results.

\section{H1 Predicts the Deterioration of Accuracy in Multi-Operand Addition}
\label{sec:multi_fail}

As shown in the last section, \textbf{H1} is a good approximator for LLM behaviour on two-operand addition: In the majority of cases, a lookahead of one digit is sufficient to accurately determine the value of the carry bit affecting $s_2$. With a look-ahead of one digit, \textbf{H1} predicts a failure of the generation of $s_2$, if and only if the value of $s_1$ does not suffice to determine the value of the carry bit. 
In two-operand addition in base 10, this is the case if and only if $t_1 = 9$.
We now show that \textbf{H1} can also account for model performance on \textit{multi}-operand addition. 

\subsection{Multi-Operand Performance Predicted by H1}

The possible value of a carry increases with increasing numbers of operands. 
For instance in 4-operand addition ($k=4$) the maximal value of a carry is $3$:
\[c_{max}(4) = \left\lfloor \frac{\sum_{j=1}^4 9}{10} \right\rfloor = 3\] 

Therefore the carry heuristic \(c_{i}^{h}\) is unreliable in 4-operand addition whenever \(t_{i-1} =\sum_{j=1}^k n_{j,i-1} \in \{7, 8, 9, 17, 18, 19, 27, 28, 29\} \).

Put simply, because the value of the carry can be larger for more operands,  \textbf{the proportion of values of \(s_1\) for which the heuristic is insufficient (with its lookahead of one) increases with an increasing number of operands}. 

Consider an example in which the heuristic fails in 4-operand addition for clarification (see Figure \ref{fig:carry_4_op_fail} in Appendix \ref{appendix:C}): 

\noindent\textbf{186 + 261 + 198 + 256.}
\[
\begin{split}
    t_{1} =8 + 6 + 9 + 5 = 28\\
    c_{2}^{h} \in \{ \left\lfloor \frac{c_{min} + 28}{10} \right\rfloor,\\
    \left\lfloor \frac{c_{max} + 28}{10} \right\rfloor \}
\end{split}
\]

with \(c_{max} = 3\)
\[c_{2}^{h} \in \{ \left\lfloor \frac{28}{10} \right\rfloor, \left\lfloor \frac{31}{10} \right\rfloor \} = \{2, 3\}\]
therefore $c_{2}^{h}$ is chosen uniformly at random between $2$ and $3$.
The heuristic thus fails in solving \textbf{186 + 261 + 198 + 256} with a chance of 50\%.

For 4-operand addition, there are 37 possible sums for the second digits (ranging from 0 to 36). In 28 out of these 37 cases, the heuristic reliably determines the correct carry bit. However, when \(t_1 \in \{7, 8, 9, 17, 18, 19, 27, 28, 29\}\), the heuristic must randomly choose between two possible carry values, leading to a 50\% chance of selecting the correct one. This results in an overall accuracy of:
\[\frac{28\times1.0 + 9 \times 0.5}{37} = 0.878\]
Thus, the heuristic only achieves 88\% accuracy in correctly predicting the first result digit $s_2$ in 4-operand addition, compared to the 97\% accuracy in two-operand addition. 
In Appendix \ref{appendix:D}, we provide exact values for $s_2$ accuracy as predicted by \textbf{H1}, for addition tasks between 2 and 11 operands.

\begin{figure}[t]
    \centering
    \includegraphics[width=0.45\textwidth]{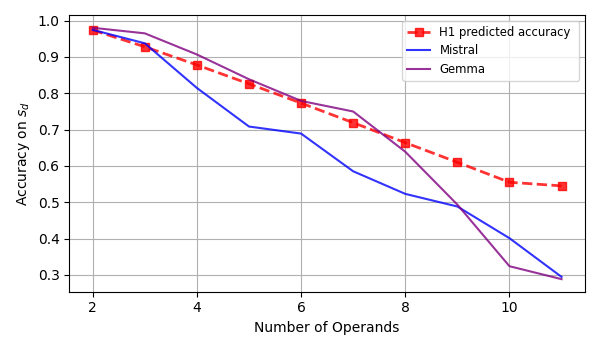}
    \caption{Accuracy of first generated result digit $s_d$ in one-shot multi-operand addition for Mistral and Gemma, compared to the expected accuracy based on \textbf{H1}.}
    \label{fig:multi_op_accuracy}
    \vspace{-0.2cm}
\end{figure}

\subsection{Empricial Evidence on Multi-Operand Addition}
Intuitively, according to \textbf{H1}, Mistral and Gemma with their one-digit tokenization should fail at multi-operand addition at a certain rate: The amount of instances in which a lookahead of one digit is sufficient to accurately predict $s_i$ gets smaller and smaller because the carry bit value can get larger and larger for multiple operands. 
We test if \textbf{H1} holds in predicting the first generated digit $s_d$ in Mistral and Gemma for multiple operands. We evaluate prediction accuracy on the multi-operand datasets described in Section \ref{sec:models_data}.
\textbf{H1} should provide an upper bound for the performance of LLMs\footnote{Autoregressive LLMs with single-digit tokenization of numbers.} for predicting the first result digit $s_d$.
Figure \ref{fig:multi_op_accuracy} shows that \textbf{H1} is a good predictor for the accuracy of the one-shot\footnote{Results for the zero-shot setting are in Appendix \ref{sec:appendix_E}.} generation of the first result digit $s_d$ by Mistral and Gemma. We take this as further evidence that these LLMs make use of \textbf{H1}.
\section{Multi-Digit Tokenization Models Employ the Same Heuristic}
\label{sec:llama}
While \citet{levy2024language} demonstrate that all LLMs, regardless of the tokenization strategy, internally represent numbers as individual digits, it remained unclear whether models with multi-digit tokenization also rely on a one-digit lookahead when generating addition results. In this section, we show that perhaps surprisingly multi-digit tokenization models, such as Llama-3, also employ a lookahead of one \textbf{digit} when predicting carry bits. 
To show this, we design 3 controlled datasets that force the multi-digit tokenization model Llama-3 to generate results across multiple tokens. 

\paragraph{Experimental Setup.}
To examine whether Llama-3 employs a one-digit lookahead, we use six-digit numbers in two-operand addition (e.g., ``231234 + 124514 = ''), where each operand is tokenized into two three-digit tokens by the model's tokenizer, such as: [`` 231'',`` 234'', `` +'', `` 124'', `` 514'', `` =''] and the result is generated as two triple-digit tokens as well, in this example [`` 355'', `` 748'']. The first generated triple-digit token $s_5 s_4 s_3$ corresponds to digit base positions $10^5$, $10^4$, and $10^3$. If Llama-3 did employ \textbf{H1} it would look ahead to digit position $10^2$, but ignore digit positions $10^1$ and $10^0$, as they fall outside the lookahead window.

\paragraph{Carry Scenarios.}
We evaluate model behavior in three datasets with six-digit operands (ranging from 100,000 to 899,999) and results between 200,000 and 999,999. We use a zero-shot prompt template. Each dataset consist of 100 samples:
\begin{itemize}
    \setlength{\itemsep}{0.1pt}
    \setlength{\parskip}{0pt}
    \setlength{\parsep}{0.5pt}
    \item \textbf{DS6: No carry.} The addition does not produce any carry and no digits sum to 9.  (e.g., \(111,234 + 111,514 = 222,748\)).
    \item \textbf{DS7: Direct carry in position \(10^2\).} A carry is generated at \(10^2\) and directly affects \(10^3\) (e.g., \(111,721 + 111,435 = 223,156\)). 
    \item \textbf{DS8: Cascading carry from \(10^1\) to \(10^3\).} A carry originates at \(10^1\), cascades to \(10^2\) and then affects \(10^3\) (e.g., \(111,382 + 111,634 = 223,016\)).
\end{itemize}

\paragraph{Expected Outcomes.}
If Llama-3 employs \textbf{H1}, we expect that DS6 should be easy, as no carry propagation is required. DS7 should also be easy, since the carry affecting \(10^3\) is within the one-digit lookahead window. DS8 in contrast should be challenging, as the carry originates from \(10^1\), from beyond the model’s lookahead range. We expect a lower accuracy in generating \(10^3\), the result digit that is affected by the potentially inaccurate carry.

\begin{figure}[t]
    \centering
    \includegraphics[width=0.4\textwidth]{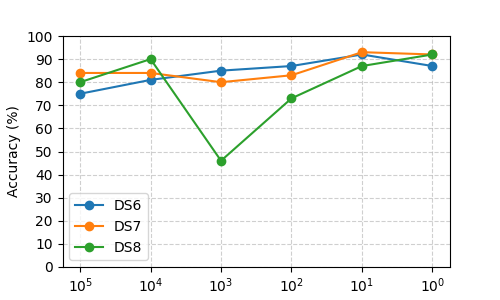} 
    \caption{Per-digit generation accuracy of Llama on datasets DS6-DS8. Each dataset represents a different carry scenario.} 
    \label{fig:llama_carry_scenarios}
\end{figure}

\paragraph{Results.}
Figure \ref{fig:llama_carry_scenarios} shows that Llama-3 exhibits the expected pattern predicted by \textbf{H1}. The sharp drop in accuracy in dataset DS8 on digit \(10^3\) provides evidence that Llama-3, regardless of its multi-digit tokenization strategy, relies on the same one-digit lookahead for solving addition left to right. 
\section{Related Work}
\label{sec:related-work}
Recent work has benchmarked the arithmetic capabilities of LLMs using text-based evaluations and handcrafted tests \cite{yuan2023well,lightman2023lets,NEURIPS2023_58168e8a,zhuang2023efficiently}. Numerous studies consistently show that LLMs struggle with arithmetic tasks \cite{nogueira2021investigatinglimitationstransformerssimple, qian2022limitationslanguagemodelsarithmetic, dziri2023faithfatelimitstransformers, yu2024metamathbootstrapmathematicalquestions}. 

\citet{zhou2023algorithmstransformerslearnstudy} and \citet{zhou2024transformersachievelengthgeneralization} examine transformers' ability to learn algorithmic procedures and find challenges in length generalization \cite{anil2022exploringlengthgeneralizationlarge}. Similarly, \citet{xiao2024theorylengthgeneralizationlearning} propose a theoretical explanation for LLMs' difficulties with length generalization in arithmetic. \citet{gambardella2024language} find that LLMs can reliably predict the first digit in multiplication but struggle with subsequent digits.

The focus of research has recently shifted from mere benchmarking of LLMs to trying to understand \textit{why} LLMs struggle with arithmetic reasoning. Using circuit analysis, \citet{stolfo_mechanistic_2023} and \citet{hanna2023doesgpt2computegreaterthan} explore internal processing in arithmetic tasks, while \citet{nikankin2024arithmetic} reveal that LLMs use a variety of heuristics managed by identifiable circuits and neurons. In contrast, \citet{deng2024language} argue that LLMs rely on symbolic pattern recognition rather than true numerical computation. Recently, \citet{kantamneni2025languagemodelsusetrigonometry} showed that LLMs represent numbers as generalized helixes and perform addition using a “Clock” algorithm \cite{nanda_progress_2023}.

Related work has also examined how LLMs encode numbers. \citet{levy2024language} demonstrate that numbers are represented digit-by-digit, extending \citet{gould2023successor}, who find that LLMs encode numeric values modulo 10. \citet{zhu-etal-2025-language} suggest that numbers are encoded linearly, while \citet{marjieh2025number} indicate that number representations can blend string-like and numerical forms.

Another line of research explores how tokenization influences arithmetic capabilities. \citet{lee2024digitstodecisions} show that single-digit tokenization outperforms other methods in simple arithmetic tasks. \citet{singh2024tokenization} highlight that right-to-left (R2L) tokenization—where tokens are right-aligned—improves arithmetic performance. 
Additionally, the role of embeddings and positional encodings is emphasized by \citet{mcleish2024transformers}, who demonstrate that suitable embeddings enable transformers to learn arithmetic, and by \citet{shen2023positionaldescriptionmatterstransformers}, who show that positional encoding improves arithmetic performance.

\section{Conclusion}

Our study shows that LLMs, regardless of their numeric tokenization strategy, rely on a simple one-digit lookahead heuristic for anticipating carries when performing addition tasks. While this strategy is fairly effective for two-operand additions, it fails in the multi-operand additions due to the increasingly unpredictable value of cascading carry bits. Through probing experiments and targeted evaluations of digit-wise result accuracy, we demonstrate that model accuracy deteriorates precisely at the rate the heuristic predicts. 

These findings highlight an inherent weakness in current LLMs that prevents them from robustly generalizing to more complex arithmetic tasks. 

Our work contributes to a broader understanding of LLM limitations in arithmetic reasoning and highlights increasing LLMs' lookahead as a promising approach to enhancing their ability to handle complex numerical tasks.

\section*{Limitations}
Our work highlights limited lookahead as a key challenge for LLMs when adding multiple numbers. However, it remains unclear whether this limitation extends to other arithmetic operations, such as subtraction. Additionally, we cannot determine whether the limited lookahead is a heuristic explicitly learned for arithmetic tasks, or if it could also affect general language generation tasks as thus hinder performance of other tasks that require long-range dependencies. Future work should explore the depth of lookahead in tasks beyond arithmetic.

While the lookahead heuristic offers a straightforward explanation for the upper performance limit of LLMs on addition, it does not fully account for why LLMs still somewhat underperform relative to the heuristic in addition tasks with many operands (e.g., adding 8–11 numbers). We suspect this discrepancy may be related to limited training exposure to these many-operand addition tasks, but further investigation is needed to confirm this.

Our work also does not address whether larger models within the same family (e.g., 70B parameter models) exhibit a deeper lookahead. Future studies should examine whether scaling model size leads to improved performance by enabling a deeper lookahead.

Finally, we do not tackle methods to overcome the shallow lookahead. Future work should investigate whether targeted training on tasks requiring deeper lookahead can encourage models to deepen their lookahead.


\section*{Acknowledgements}
We thank Patrick Schramowski for his helpful feedback on the paper draft. This work has been supported by the German Ministry of Education and Research (BMBF) as part of the project TRAILS (01IW24005).

\bibliography{custom, custom1}

\begin{thebibliography}{33}
\providecommand{\natexlab}[1]{#1}

\bibitem[{AI@Meta(2024)}]{llama3modelcard}
AI@Meta. 2024.
\newblock \href {https://github.com/meta-llama/llama3/blob/main/MODEL_CARD.md} {Llama 3 model card}.

\bibitem[{Anil et~al.(2022)Anil, Wu, Andreassen, Lewkowycz, Misra, Ramasesh, Slone, Gur-Ari, Dyer, and Neyshabur}]{anil2022exploringlengthgeneralizationlarge}
Cem Anil, Yuhuai Wu, Anders Andreassen, Aitor Lewkowycz, Vedant Misra, Vinay Ramasesh, Ambrose Slone, Guy Gur-Ari, Ethan Dyer, and Behnam Neyshabur. 2022.
\newblock \href {https://arxiv.org/abs/2207.04901} {Exploring length generalization in large language models}.
\newblock \emph{Preprint}, arXiv:2207.04901.

\bibitem[{Bai et~al.(2023)Bai, Bai, Chu, Cui, Dang, Deng, Fan, Ge, Han, Huang et~al.}]{bai2023qwen}
Jinze Bai, Shuai Bai, Yunfei Chu, Zeyu Cui, Kai Dang, Xiaodong Deng, Yang Fan, Wenbin Ge, Yu~Han, Fei Huang, et~al. 2023.
\newblock Qwen technical report.
\newblock \emph{arXiv preprint arXiv:2309.16609}.

\bibitem[{Deng et~al.(2024)Deng, Li, Xie, Chang, and Chen}]{deng2024language}
Chunyuan Deng, Zhiqi Li, Roy Xie, Ruidi Chang, and Hanjie Chen. 2024.
\newblock Language models are symbolic learners in arithmetic.
\newblock \emph{arXiv preprint arXiv:2410.15580}.

\bibitem[{Dziri et~al.(2023)Dziri, Lu, Sclar, Li, Jiang, Lin, West, Bhagavatula, Bras, Hwang, Sanyal, Welleck, Ren, Ettinger, Harchaoui, and Choi}]{dziri2023faithfatelimitstransformers}
Nouha Dziri, Ximing Lu, Melanie Sclar, Xiang~Lorraine Li, Liwei Jiang, Bill~Yuchen Lin, Peter West, Chandra Bhagavatula, Ronan~Le Bras, Jena~D. Hwang, Soumya Sanyal, Sean Welleck, Xiang Ren, Allyson Ettinger, Zaid Harchaoui, and Yejin Choi. 2023.
\newblock \href {https://arxiv.org/abs/2305.18654} {Faith and fate: Limits of transformers on compositionality}.
\newblock \emph{Preprint}, arXiv:2305.18654.

\bibitem[{Frieder et~al.(2023)Frieder, Pinchetti, , Griffiths, Salvatori, Lukasiewicz, Petersen, and Berner}]{NEURIPS2023_58168e8a}
Simon Frieder, Luca Pinchetti, , Ryan-Rhys Griffiths, Tommaso Salvatori, Thomas Lukasiewicz, Philipp Petersen, and Julius Berner. 2023.
\newblock \href {https://proceedings.neurips.cc/paper_files/paper/2023/file/58168e8a92994655d6da3939e7cc0918-Paper-Datasets_and_Benchmarks.pdf} {Mathematical capabilities of chatgpt}.
\newblock In \emph{Advances in Neural Information Processing Systems}, volume~36, pages 27699--27744. Curran Associates, Inc.

\bibitem[{Gambardella et~al.(2024)Gambardella, Iwasawa, and Matsuo}]{gambardella2024language}
Andrew Gambardella, Yusuke Iwasawa, and Yutaka Matsuo. 2024.
\newblock Language models do hard arithmetic tasks easily and hardly do easy arithmetic tasks.
\newblock \emph{arXiv preprint arXiv:2406.02356}.

\bibitem[{Garreth~Lee and Wolf(2024)}]{lee2024digitstodecisions}
Leandro von~Werra Garreth~Lee, Guilherme~Penedo and Thomas Wolf. 2024.
\newblock \href {https://huggingface.co/spaces/huggingface/number-tokenization-blog} {From digits to decisions: How tokenization impacts arithmetic in llms}.

\bibitem[{Gould et~al.(2023)Gould, Ong, Ogden, and Conmy}]{gould2023successor}
Rhys Gould, Euan Ong, George Ogden, and Arthur Conmy. 2023.
\newblock Successor heads: Recurring, interpretable attention heads in the wild.
\newblock \emph{arXiv preprint arXiv:2312.09230}.

\bibitem[{Grattafiori et~al.(2024)Grattafiori, Dubey, Jauhri, Pandey, Kadian, Al-Dahle, Letman, Mathur, Schelten, Vaughan, Yang, Fan, Goyal, Hartshorn, Yang, Mitra, Sravankumar, Korenev, Hinsvark, Rao, Zhang, Rodriguez, Gregerson, Spataru, Roziere, Biron, Tang, Chern, Caucheteux, Nayak, Bi, Marra, McConnell, Keller, Touret, Wu, Wong, Ferrer, Nikolaidis, Allonsius, Song, Pintz, Livshits, Wyatt, Esiobu, Choudhary, Mahajan, Garcia-Olano, Perino, Hupkes, Lakomkin, AlBadawy, Lobanova, Dinan, Smith, Radenovic, Guzmán, Zhang, Synnaeve, Lee, Anderson, Thattai, Nail, Mialon, Pang, Cucurell, Nguyen, Korevaar, Xu, Touvron, Zarov, Ibarra, Kloumann, Misra, Evtimov, Zhang, Copet, Lee, Geffert, Vranes, Park, Mahadeokar, Shah, van~der Linde, Billock, Hong, Lee, Fu, Chi, Huang, Liu, Wang, Yu, Bitton, Spisak, Park, Rocca, Johnstun, Saxe, Jia, Alwala, Prasad, Upasani, Plawiak, Li, Heafield, Stone, El-Arini, Iyer, Malik, Chiu, Bhalla, Lakhotia, Rantala-Yeary, van~der Maaten, Chen, Tan, Jenkins, Martin, Madaan, Malo, Blecher,
  Landzaat, de~Oliveira, Muzzi, Pasupuleti, Singh, Paluri, Kardas, Tsimpoukelli, Oldham, Rita, Pavlova, Kambadur, Lewis, Si, Singh, Hassan, Goyal, Torabi, Bashlykov, Bogoychev, Chatterji, Zhang, Duchenne, Çelebi, Alrassy, Zhang, Li, Vasic, Weng, Bhargava, Dubal, Krishnan, Koura, Xu, He, Dong, Srinivasan, Ganapathy, Calderer, Cabral, Stojnic, Raileanu, Maheswari, Girdhar, Patel, Sauvestre, Polidoro, Sumbaly, Taylor, Silva, Hou, Wang, Hosseini, Chennabasappa, Singh, Bell, Kim, Edunov, Nie, Narang, Raparthy, Shen, Wan, Bhosale, Zhang, Vandenhende, Batra, Whitman, Sootla, Collot, Gururangan, Borodinsky, Herman, Fowler, Sheasha, Georgiou, Scialom, Speckbacher, Mihaylov, Xiao, Karn, Goswami, Gupta, Ramanathan, Kerkez, Gonguet, Do, Vogeti, Albiero, Petrovic, Chu, Xiong, Fu, Meers, Martinet, Wang, Wang, Tan, Xia, Xie, Jia, Wang, Goldschlag, Gaur, Babaei, Wen, Song, Zhang, Li, Mao, Coudert, Yan, Chen, Papakipos, Singh, Srivastava, Jain, Kelsey, Shajnfeld, Gangidi, Victoria, Goldstand, Menon, Sharma, Boesenberg,
  Baevski, Feinstein, Kallet, Sangani, Teo, Yunus, Lupu, Alvarado, Caples, Gu, Ho, Poulton, Ryan, Ramchandani, Dong, Franco, Goyal, Saraf, Chowdhury, Gabriel, Bharambe, Eisenman, Yazdan, James, Maurer, Leonhardi, Huang, Loyd, Paola, Paranjape, Liu, Wu, Ni, Hancock, Wasti, Spence, Stojkovic, Gamido, Montalvo, Parker, Burton, Mejia, Liu, Wang, Kim, Zhou, Hu, Chu, Cai, Tindal, Feichtenhofer, Gao, Civin, Beaty, Kreymer, Li, Adkins, Xu, Testuggine, David, Parikh, Liskovich, Foss, Wang, Le, Holland, Dowling, Jamil, Montgomery, Presani, Hahn, Wood, Le, Brinkman, Arcaute, Dunbar, Smothers, Sun, Kreuk, Tian, Kokkinos, Ozgenel, Caggioni, Kanayet, Seide, Florez, Schwarz, Badeer, Swee, Halpern, Herman, Sizov, Guangyi, Zhang, Lakshminarayanan, Inan, Shojanazeri, Zou, Wang, Zha, Habeeb, Rudolph, Suk, Aspegren, Goldman, Zhan, Damlaj, Molybog, Tufanov, Leontiadis, Veliche, Gat, Weissman, Geboski, Kohli, Lam, Asher, Gaya, Marcus, Tang, Chan, Zhen, Reizenstein, Teboul, Zhong, Jin, Yang, Cummings, Carvill, Shepard, McPhie,
  Torres, Ginsburg, Wang, Wu, U, Saxena, Khandelwal, Zand, Matosich, Veeraraghavan, Michelena, Li, Jagadeesh, Huang, Chawla, Huang, Chen, Garg, A, Silva, Bell, Zhang, Guo, Yu, Moshkovich, Wehrstedt, Khabsa, Avalani, Bhatt, Mankus, Hasson, Lennie, Reso, Groshev, Naumov, Lathi, Keneally, Liu, Seltzer, Valko, Restrepo, Patel, Vyatskov, Samvelyan, Clark, Macey, Wang, Hermoso, Metanat, Rastegari, Bansal, Santhanam, Parks, White, Bawa, Singhal, Egebo, Usunier, Mehta, Laptev, Dong, Cheng, Chernoguz, Hart, Salpekar, Kalinli, Kent, Parekh, Saab, Balaji, Rittner, Bontrager, Roux, Dollar, Zvyagina, Ratanchandani, Yuvraj, Liang, Alao, Rodriguez, Ayub, Murthy, Nayani, Mitra, Parthasarathy, Li, Hogan, Battey, Wang, Howes, Rinott, Mehta, Siby, Bondu, Datta, Chugh, Hunt, Dhillon, Sidorov, Pan, Mahajan, Verma, Yamamoto, Ramaswamy, Lindsay, Lindsay, Feng, Lin, Zha, Patil, Shankar, Zhang, Zhang, Wang, Agarwal, Sajuyigbe, Chintala, Max, Chen, Kehoe, Satterfield, Govindaprasad, Gupta, Deng, Cho, Virk, Subramanian, Choudhury,
  Goldman, Remez, Glaser, Best, Koehler, Robinson, Li, Zhang, Matthews, Chou, Shaked, Vontimitta, Ajayi, Montanez, Mohan, Kumar, Mangla, Ionescu, Poenaru, Mihailescu, Ivanov, Li, Wang, Jiang, Bouaziz, Constable, Tang, Wu, Wang, Wu, Gao, Kleinman, Chen, Hu, Jia, Qi, Li, Zhang, Zhang, Adi, Nam, Yu, Wang, Zhao, Hao, Qian, Li, He, Rait, DeVito, Rosnbrick, Wen, Yang, Zhao, and Ma}]{grattafiori2024llama3herdmodels}
Aaron Grattafiori, Abhimanyu Dubey, Abhinav Jauhri, Abhinav Pandey, Abhishek Kadian, Ahmad Al-Dahle, Aiesha Letman, Akhil Mathur, Alan Schelten, Alex Vaughan, Amy Yang, Angela Fan, Anirudh Goyal, Anthony Hartshorn, Aobo Yang, Archi Mitra, Archie Sravankumar, Artem Korenev, Arthur Hinsvark, Arun Rao, Aston Zhang, Aurelien Rodriguez, Austen Gregerson, Ava Spataru, Baptiste Roziere, Bethany Biron, Binh Tang, Bobbie Chern, Charlotte Caucheteux, Chaya Nayak, Chloe Bi, Chris Marra, Chris McConnell, Christian Keller, Christophe Touret, Chunyang Wu, Corinne Wong, Cristian~Canton Ferrer, Cyrus Nikolaidis, Damien Allonsius, Daniel Song, Danielle Pintz, Danny Livshits, Danny Wyatt, David Esiobu, Dhruv Choudhary, Dhruv Mahajan, Diego Garcia-Olano, Diego Perino, Dieuwke Hupkes, Egor Lakomkin, Ehab AlBadawy, Elina Lobanova, Emily Dinan, Eric~Michael Smith, Filip Radenovic, Francisco Guzmán, Frank Zhang, Gabriel Synnaeve, Gabrielle Lee, Georgia~Lewis Anderson, Govind Thattai, Graeme Nail, Gregoire Mialon, Guan Pang,
  Guillem Cucurell, Hailey Nguyen, Hannah Korevaar, Hu~Xu, Hugo Touvron, Iliyan Zarov, Imanol~Arrieta Ibarra, Isabel Kloumann, Ishan Misra, Ivan Evtimov, Jack Zhang, Jade Copet, Jaewon Lee, Jan Geffert, Jana Vranes, Jason Park, Jay Mahadeokar, Jeet Shah, Jelmer van~der Linde, Jennifer Billock, Jenny Hong, Jenya Lee, Jeremy Fu, Jianfeng Chi, Jianyu Huang, Jiawen Liu, Jie Wang, Jiecao Yu, Joanna Bitton, Joe Spisak, Jongsoo Park, Joseph Rocca, Joshua Johnstun, Joshua Saxe, Junteng Jia, Kalyan~Vasuden Alwala, Karthik Prasad, Kartikeya Upasani, Kate Plawiak, Ke~Li, Kenneth Heafield, Kevin Stone, Khalid El-Arini, Krithika Iyer, Kshitiz Malik, Kuenley Chiu, Kunal Bhalla, Kushal Lakhotia, Lauren Rantala-Yeary, Laurens van~der Maaten, Lawrence Chen, Liang Tan, Liz Jenkins, Louis Martin, Lovish Madaan, Lubo Malo, Lukas Blecher, Lukas Landzaat, Luke de~Oliveira, Madeline Muzzi, Mahesh Pasupuleti, Mannat Singh, Manohar Paluri, Marcin Kardas, Maria Tsimpoukelli, Mathew Oldham, Mathieu Rita, Maya Pavlova, Melanie Kambadur,
  Mike Lewis, Min Si, Mitesh~Kumar Singh, Mona Hassan, Naman Goyal, Narjes Torabi, Nikolay Bashlykov, Nikolay Bogoychev, Niladri Chatterji, Ning Zhang, Olivier Duchenne, Onur Çelebi, Patrick Alrassy, Pengchuan Zhang, Pengwei Li, Petar Vasic, Peter Weng, Prajjwal Bhargava, Pratik Dubal, Praveen Krishnan, Punit~Singh Koura, Puxin Xu, Qing He, Qingxiao Dong, Ragavan Srinivasan, Raj Ganapathy, Ramon Calderer, Ricardo~Silveira Cabral, Robert Stojnic, Roberta Raileanu, Rohan Maheswari, Rohit Girdhar, Rohit Patel, Romain Sauvestre, Ronnie Polidoro, Roshan Sumbaly, Ross Taylor, Ruan Silva, Rui Hou, Rui Wang, Saghar Hosseini, Sahana Chennabasappa, Sanjay Singh, Sean Bell, Seohyun~Sonia Kim, Sergey Edunov, Shaoliang Nie, Sharan Narang, Sharath Raparthy, Sheng Shen, Shengye Wan, Shruti Bhosale, Shun Zhang, Simon Vandenhende, Soumya Batra, Spencer Whitman, Sten Sootla, Stephane Collot, Suchin Gururangan, Sydney Borodinsky, Tamar Herman, Tara Fowler, Tarek Sheasha, Thomas Georgiou, Thomas Scialom, Tobias Speckbacher,
  Todor Mihaylov, Tong Xiao, Ujjwal Karn, Vedanuj Goswami, Vibhor Gupta, Vignesh Ramanathan, Viktor Kerkez, Vincent Gonguet, Virginie Do, Vish Vogeti, Vítor Albiero, Vladan Petrovic, Weiwei Chu, Wenhan Xiong, Wenyin Fu, Whitney Meers, Xavier Martinet, Xiaodong Wang, Xiaofang Wang, Xiaoqing~Ellen Tan, Xide Xia, Xinfeng Xie, Xuchao Jia, Xuewei Wang, Yaelle Goldschlag, Yashesh Gaur, Yasmine Babaei, Yi~Wen, Yiwen Song, Yuchen Zhang, Yue Li, Yuning Mao, Zacharie~Delpierre Coudert, Zheng Yan, Zhengxing Chen, Zoe Papakipos, Aaditya Singh, Aayushi Srivastava, Abha Jain, Adam Kelsey, Adam Shajnfeld, Adithya Gangidi, Adolfo Victoria, Ahuva Goldstand, Ajay Menon, Ajay Sharma, Alex Boesenberg, Alexei Baevski, Allie Feinstein, Amanda Kallet, Amit Sangani, Amos Teo, Anam Yunus, Andrei Lupu, Andres Alvarado, Andrew Caples, Andrew Gu, Andrew Ho, Andrew Poulton, Andrew Ryan, Ankit Ramchandani, Annie Dong, Annie Franco, Anuj Goyal, Aparajita Saraf, Arkabandhu Chowdhury, Ashley Gabriel, Ashwin Bharambe, Assaf Eisenman, Azadeh
  Yazdan, Beau James, Ben Maurer, Benjamin Leonhardi, Bernie Huang, Beth Loyd, Beto~De Paola, Bhargavi Paranjape, Bing Liu, Bo~Wu, Boyu Ni, Braden Hancock, Bram Wasti, Brandon Spence, Brani Stojkovic, Brian Gamido, Britt Montalvo, Carl Parker, Carly Burton, Catalina Mejia, Ce~Liu, Changhan Wang, Changkyu Kim, Chao Zhou, Chester Hu, Ching-Hsiang Chu, Chris Cai, Chris Tindal, Christoph Feichtenhofer, Cynthia Gao, Damon Civin, Dana Beaty, Daniel Kreymer, Daniel Li, David Adkins, David Xu, Davide Testuggine, Delia David, Devi Parikh, Diana Liskovich, Didem Foss, Dingkang Wang, Duc Le, Dustin Holland, Edward Dowling, Eissa Jamil, Elaine Montgomery, Eleonora Presani, Emily Hahn, Emily Wood, Eric-Tuan Le, Erik Brinkman, Esteban Arcaute, Evan Dunbar, Evan Smothers, Fei Sun, Felix Kreuk, Feng Tian, Filippos Kokkinos, Firat Ozgenel, Francesco Caggioni, Frank Kanayet, Frank Seide, Gabriela~Medina Florez, Gabriella Schwarz, Gada Badeer, Georgia Swee, Gil Halpern, Grant Herman, Grigory Sizov, Guangyi, Zhang, Guna
  Lakshminarayanan, Hakan Inan, Hamid Shojanazeri, Han Zou, Hannah Wang, Hanwen Zha, Haroun Habeeb, Harrison Rudolph, Helen Suk, Henry Aspegren, Hunter Goldman, Hongyuan Zhan, Ibrahim Damlaj, Igor Molybog, Igor Tufanov, Ilias Leontiadis, Irina-Elena Veliche, Itai Gat, Jake Weissman, James Geboski, James Kohli, Janice Lam, Japhet Asher, Jean-Baptiste Gaya, Jeff Marcus, Jeff Tang, Jennifer Chan, Jenny Zhen, Jeremy Reizenstein, Jeremy Teboul, Jessica Zhong, Jian Jin, Jingyi Yang, Joe Cummings, Jon Carvill, Jon Shepard, Jonathan McPhie, Jonathan Torres, Josh Ginsburg, Junjie Wang, Kai Wu, Kam~Hou U, Karan Saxena, Kartikay Khandelwal, Katayoun Zand, Kathy Matosich, Kaushik Veeraraghavan, Kelly Michelena, Keqian Li, Kiran Jagadeesh, Kun Huang, Kunal Chawla, Kyle Huang, Lailin Chen, Lakshya Garg, Lavender A, Leandro Silva, Lee Bell, Lei Zhang, Liangpeng Guo, Licheng Yu, Liron Moshkovich, Luca Wehrstedt, Madian Khabsa, Manav Avalani, Manish Bhatt, Martynas Mankus, Matan Hasson, Matthew Lennie, Matthias Reso, Maxim
  Groshev, Maxim Naumov, Maya Lathi, Meghan Keneally, Miao Liu, Michael~L. Seltzer, Michal Valko, Michelle Restrepo, Mihir Patel, Mik Vyatskov, Mikayel Samvelyan, Mike Clark, Mike Macey, Mike Wang, Miquel~Jubert Hermoso, Mo~Metanat, Mohammad Rastegari, Munish Bansal, Nandhini Santhanam, Natascha Parks, Natasha White, Navyata Bawa, Nayan Singhal, Nick Egebo, Nicolas Usunier, Nikhil Mehta, Nikolay~Pavlovich Laptev, Ning Dong, Norman Cheng, Oleg Chernoguz, Olivia Hart, Omkar Salpekar, Ozlem Kalinli, Parkin Kent, Parth Parekh, Paul Saab, Pavan Balaji, Pedro Rittner, Philip Bontrager, Pierre Roux, Piotr Dollar, Polina Zvyagina, Prashant Ratanchandani, Pritish Yuvraj, Qian Liang, Rachad Alao, Rachel Rodriguez, Rafi Ayub, Raghotham Murthy, Raghu Nayani, Rahul Mitra, Rangaprabhu Parthasarathy, Raymond Li, Rebekkah Hogan, Robin Battey, Rocky Wang, Russ Howes, Ruty Rinott, Sachin Mehta, Sachin Siby, Sai~Jayesh Bondu, Samyak Datta, Sara Chugh, Sara Hunt, Sargun Dhillon, Sasha Sidorov, Satadru Pan, Saurabh Mahajan,
  Saurabh Verma, Seiji Yamamoto, Sharadh Ramaswamy, Shaun Lindsay, Shaun Lindsay, Sheng Feng, Shenghao Lin, Shengxin~Cindy Zha, Shishir Patil, Shiva Shankar, Shuqiang Zhang, Shuqiang Zhang, Sinong Wang, Sneha Agarwal, Soji Sajuyigbe, Soumith Chintala, Stephanie Max, Stephen Chen, Steve Kehoe, Steve Satterfield, Sudarshan Govindaprasad, Sumit Gupta, Summer Deng, Sungmin Cho, Sunny Virk, Suraj Subramanian, Sy~Choudhury, Sydney Goldman, Tal Remez, Tamar Glaser, Tamara Best, Thilo Koehler, Thomas Robinson, Tianhe Li, Tianjun Zhang, Tim Matthews, Timothy Chou, Tzook Shaked, Varun Vontimitta, Victoria Ajayi, Victoria Montanez, Vijai Mohan, Vinay~Satish Kumar, Vishal Mangla, Vlad Ionescu, Vlad Poenaru, Vlad~Tiberiu Mihailescu, Vladimir Ivanov, Wei Li, Wenchen Wang, Wenwen Jiang, Wes Bouaziz, Will Constable, Xiaocheng Tang, Xiaojian Wu, Xiaolan Wang, Xilun Wu, Xinbo Gao, Yaniv Kleinman, Yanjun Chen, Ye~Hu, Ye~Jia, Ye~Qi, Yenda Li, Yilin Zhang, Ying Zhang, Yossi Adi, Youngjin Nam, Yu, Wang, Yu~Zhao, Yuchen Hao, Yundi
  Qian, Yunlu Li, Yuzi He, Zach Rait, Zachary DeVito, Zef Rosnbrick, Zhaoduo Wen, Zhenyu Yang, Zhiwei Zhao, and Zhiyu Ma. 2024.
\newblock \href {https://arxiv.org/abs/2407.21783} {The llama 3 herd of models}.
\newblock \emph{Preprint}, arXiv:2407.21783.

\bibitem[{Guo et~al.(2025)Guo, Yang, Zhang, Song, Zhang, Xu, Zhu, Ma, Wang, Bi et~al.}]{guo2025deepseek}
Daya Guo, Dejian Yang, Haowei Zhang, Junxiao Song, Ruoyu Zhang, Runxin Xu, Qihao Zhu, Shirong Ma, Peiyi Wang, Xiao Bi, et~al. 2025.
\newblock Deepseek-r1: Incentivizing reasoning capability in llms via reinforcement learning.
\newblock \emph{arXiv preprint arXiv:2501.12948}.

\bibitem[{Hanna et~al.(2023)Hanna, Liu, and Variengien}]{hanna2023doesgpt2computegreaterthan}
Michael Hanna, Ollie Liu, and Alexandre Variengien. 2023.
\newblock \href {https://arxiv.org/abs/2305.00586} {How does gpt-2 compute greater-than?: Interpreting mathematical abilities in a pre-trained language model}.
\newblock \emph{Preprint}, arXiv:2305.00586.

\bibitem[{Jiang et~al.(2023)Jiang, Sablayrolles, Mensch, Bamford, Chaplot, Casas, Bressand, Lengyel, Lample, Saulnier, Lavaud, Lachaux, Stock, Scao, Lavril, Wang, Lacroix, and Sayed}]{jiang_mistral_2023}
Albert~Q. Jiang, Alexandre Sablayrolles, Arthur Mensch, Chris Bamford, Devendra~Singh Chaplot, Diego de~las Casas, Florian Bressand, Gianna Lengyel, Guillaume Lample, Lucile Saulnier, Lélio~Renard Lavaud, Marie-Anne Lachaux, Pierre Stock, Teven~Le Scao, Thibaut Lavril, Thomas Wang, Timothée Lacroix, and William~El Sayed. 2023.
\newblock \href {https://doi.org/10.48550/arXiv.2310.06825} {Mistral {7B}}.
\newblock \emph{arXiv preprint}.
\newblock ArXiv:2310.06825 [cs].

\bibitem[{Kantamneni and Tegmark(2025)}]{kantamneni2025languagemodelsusetrigonometry}
Subhash Kantamneni and Max Tegmark. 2025.
\newblock \href {https://arxiv.org/abs/2502.00873} {Language models use trigonometry to do addition}.
\newblock \emph{Preprint}, arXiv:2502.00873.

\bibitem[{Levy and Geva(2024)}]{levy2024language}
Amit~Arnold Levy and Mor Geva. 2024.
\newblock Language models encode numbers using digit representations in base 10.
\newblock \emph{arXiv preprint arXiv:2410.11781}.

\bibitem[{Lightman et~al.(2023)Lightman, Kosaraju, Burda, Edwards, Baker, Lee, Leike, Schulman, Sutskever, and Cobbe}]{lightman2023lets}
Hunter Lightman, Vineet Kosaraju, Yura Burda, Harri Edwards, Bowen Baker, Teddy Lee, Jan Leike, John Schulman, Ilya Sutskever, and Karl Cobbe. 2023.
\newblock Let's verify step by step.
\newblock \emph{arXiv preprint arXiv:2305.20050}.

\bibitem[{Marjieh et~al.(2025)Marjieh, Veselovsky, Griffiths, and Sucholutsky}]{marjieh2025number}
Raja Marjieh, Veniamin Veselovsky, Thomas~L Griffiths, and Ilia Sucholutsky. 2025.
\newblock What is a number, that a large language model may know it?
\newblock \emph{arXiv preprint arXiv:2502.01540}.

\bibitem[{McLeish et~al.(2024)McLeish, Bansal, Stein, Jain, Kirchenbauer, Bartoldson, Kailkhura, Bhatele, Geiping, Schwarzschild, and Goldstein}]{mcleish2024transformers}
Sean~Michael McLeish, Arpit Bansal, Alex Stein, Neel Jain, John Kirchenbauer, Brian~R. Bartoldson, Bhavya Kailkhura, Abhinav Bhatele, Jonas Geiping, Avi Schwarzschild, and Tom Goldstein. 2024.
\newblock \href {https://openreview.net/forum?id=KD9pZCuOVz} {Transformers can do arithmetic with the right embeddings}.
\newblock In \emph{ICML 2024 Workshop on LLMs and Cognition}.

\bibitem[{Nanda et~al.(2023)Nanda, Chan, Lieberum, Smith, and Steinhardt}]{nanda_progress_2023}
Neel Nanda, Lawrence Chan, Tom Lieberum, Jess Smith, and Jacob Steinhardt. 2023.
\newblock \href {https://doi.org/10.48550/arXiv.2301.05217} {Progress measures for grokking via mechanistic interpretability}.
\newblock \emph{arXiv preprint}.
\newblock ArXiv:2301.05217 [cs].

\bibitem[{Nikankin et~al.(2024)Nikankin, Reusch, Mueller, and Belinkov}]{nikankin2024arithmetic}
Yaniv Nikankin, Anja Reusch, Aaron Mueller, and Yonatan Belinkov. 2024.
\newblock Arithmetic without algorithms: Language models solve math with a bag of heuristics.
\newblock \emph{arXiv preprint arXiv:2410.21272}.

\bibitem[{Nogueira et~al.(2021)Nogueira, Jiang, and Lin}]{nogueira2021investigatinglimitationstransformerssimple}
Rodrigo Nogueira, Zhiying Jiang, and Jimmy Lin. 2021.
\newblock \href {https://arxiv.org/abs/2102.13019} {Investigating the limitations of transformers with simple arithmetic tasks}.
\newblock \emph{Preprint}, arXiv:2102.13019.

\bibitem[{Qian et~al.(2022)Qian, Wang, Li, Li, and Yan}]{qian2022limitationslanguagemodelsarithmetic}
Jing Qian, Hong Wang, Zekun Li, Shiyang Li, and Xifeng Yan. 2022.
\newblock \href {https://arxiv.org/abs/2208.05051} {Limitations of language models in arithmetic and symbolic induction}.
\newblock \emph{Preprint}, arXiv:2208.05051.

\bibitem[{Shen et~al.(2023)Shen, Bubeck, Eldan, Lee, Li, and Zhang}]{shen2023positionaldescriptionmatterstransformers}
Ruoqi Shen, Sébastien Bubeck, Ronen Eldan, Yin~Tat Lee, Yuanzhi Li, and Yi~Zhang. 2023.
\newblock \href {https://arxiv.org/abs/2311.14737} {Positional description matters for transformers arithmetic}.
\newblock \emph{Preprint}, arXiv:2311.14737.

\bibitem[{Singh and Strouse(2024)}]{singh2024tokenization}
Aaditya~K Singh and DJ~Strouse. 2024.
\newblock Tokenization counts: the impact of tokenization on arithmetic in frontier llms.
\newblock \emph{arXiv preprint arXiv:2402.14903}.

\bibitem[{Stolfo et~al.(2023)Stolfo, Belinkov, and Sachan}]{stolfo_mechanistic_2023}
Alessandro Stolfo, Yonatan Belinkov, and Mrinmaya Sachan. 2023.
\newblock \href {http://arxiv.org/abs/2305.15054} {A {Mechanistic} {Interpretation} of {Arithmetic} {Reasoning} in {Language} {Models} using {Causal} {Mediation} {Analysis}}.
\newblock \emph{arXiv preprint}.
\newblock ArXiv:2305.15054 [cs].

\bibitem[{Team et~al.(2024)Team, Mesnard, Hardin, Dadashi, Bhupatiraju, Pathak, Sifre, Rivière, Kale, Love, Tafti, Hussenot, Sessa, Chowdhery, Roberts, Barua, Botev, Castro-Ros, Slone, Héliou, Tacchetti, Bulanova, Paterson, Tsai, Shahriari, Lan, Choquette-Choo, Crepy, Cer, Ippolito, Reid, Buchatskaya, Ni, Noland, Yan, Tucker, Muraru, Rozhdestvenskiy, Michalewski, Tenney, Grishchenko, Austin, Keeling, Labanowski, Lespiau, Stanway, Brennan, Chen, Ferret, Chiu, Mao-Jones, Lee, Yu, Millican, Sjoesund, Lee, Dixon, Reid, Mikuła, Wirth, Sharman, Chinaev, Thain, Bachem, Chang, Wahltinez, Bailey, Michel, Yotov, Chaabouni, Comanescu, Jana, Anil, McIlroy, Liu, Mullins, Smith, Borgeaud, Girgin, Douglas, Pandya, Shakeri, De, Klimenko, Hennigan, Feinberg, Stokowiec, hui Chen, Ahmed, Gong, Warkentin, Peran, Giang, Farabet, Vinyals, Dean, Kavukcuoglu, Hassabis, Ghahramani, Eck, Barral, Pereira, Collins, Joulin, Fiedel, Senter, Andreev, and Kenealy}]{gemmateam2024gemmaopenmodelsbased}
Gemma Team, Thomas Mesnard, Cassidy Hardin, Robert Dadashi, Surya Bhupatiraju, Shreya Pathak, Laurent Sifre, Morgane Rivière, Mihir~Sanjay Kale, Juliette Love, Pouya Tafti, Léonard Hussenot, Pier~Giuseppe Sessa, Aakanksha Chowdhery, Adam Roberts, Aditya Barua, Alex Botev, Alex Castro-Ros, Ambrose Slone, Amélie Héliou, Andrea Tacchetti, Anna Bulanova, Antonia Paterson, Beth Tsai, Bobak Shahriari, Charline~Le Lan, Christopher~A. Choquette-Choo, Clément Crepy, Daniel Cer, Daphne Ippolito, David Reid, Elena Buchatskaya, Eric Ni, Eric Noland, Geng Yan, George Tucker, George-Christian Muraru, Grigory Rozhdestvenskiy, Henryk Michalewski, Ian Tenney, Ivan Grishchenko, Jacob Austin, James Keeling, Jane Labanowski, Jean-Baptiste Lespiau, Jeff Stanway, Jenny Brennan, Jeremy Chen, Johan Ferret, Justin Chiu, Justin Mao-Jones, Katherine Lee, Kathy Yu, Katie Millican, Lars~Lowe Sjoesund, Lisa Lee, Lucas Dixon, Machel Reid, Maciej Mikuła, Mateo Wirth, Michael Sharman, Nikolai Chinaev, Nithum Thain, Olivier Bachem,
  Oscar Chang, Oscar Wahltinez, Paige Bailey, Paul Michel, Petko Yotov, Rahma Chaabouni, Ramona Comanescu, Reena Jana, Rohan Anil, Ross McIlroy, Ruibo Liu, Ryan Mullins, Samuel~L Smith, Sebastian Borgeaud, Sertan Girgin, Sholto Douglas, Shree Pandya, Siamak Shakeri, Soham De, Ted Klimenko, Tom Hennigan, Vlad Feinberg, Wojciech Stokowiec, Yu~hui Chen, Zafarali Ahmed, Zhitao Gong, Tris Warkentin, Ludovic Peran, Minh Giang, Clément Farabet, Oriol Vinyals, Jeff Dean, Koray Kavukcuoglu, Demis Hassabis, Zoubin Ghahramani, Douglas Eck, Joelle Barral, Fernando Pereira, Eli Collins, Armand Joulin, Noah Fiedel, Evan Senter, Alek Andreev, and Kathleen Kenealy. 2024.
\newblock \href {https://arxiv.org/abs/2403.08295} {Gemma: Open models based on gemini research and technology}.
\newblock \emph{Preprint}, arXiv:2403.08295.

\bibitem[{Xiao and Liu(2024)}]{xiao2024theorylengthgeneralizationlearning}
Changnan Xiao and Bing Liu. 2024.
\newblock \href {https://arxiv.org/abs/2404.00560} {A theory for length generalization in learning to reason}.
\newblock \emph{Preprint}, arXiv:2404.00560.

\bibitem[{Yu et~al.(2024)Yu, Jiang, Shi, Yu, Liu, Zhang, Kwok, Li, Weller, and Liu}]{yu2024metamathbootstrapmathematicalquestions}
Longhui Yu, Weisen Jiang, Han Shi, Jincheng Yu, Zhengying Liu, Yu~Zhang, James~T. Kwok, Zhenguo Li, Adrian Weller, and Weiyang Liu. 2024.
\newblock \href {https://arxiv.org/abs/2309.12284} {Metamath: Bootstrap your own mathematical questions for large language models}.
\newblock \emph{Preprint}, arXiv:2309.12284.

\bibitem[{Yuan et~al.(2023)Yuan, Yuan, Tan, Wang, and Huang}]{yuan2023well}
Zheng Yuan, Hongyi Yuan, Chuanqi Tan, Wei Wang, and Songfang Huang. 2023.
\newblock How well do large language models perform in arithmetic tasks?
\newblock \emph{arXiv preprint arXiv:2304.02015}.

\bibitem[{Zhou et~al.(2023)Zhou, Bradley, Littwin, Razin, Saremi, Susskind, Bengio, and Nakkiran}]{zhou2023algorithmstransformerslearnstudy}
Hattie Zhou, Arwen Bradley, Etai Littwin, Noam Razin, Omid Saremi, Josh Susskind, Samy Bengio, and Preetum Nakkiran. 2023.
\newblock \href {https://arxiv.org/abs/2310.16028} {What algorithms can transformers learn? a study in length generalization}.
\newblock \emph{Preprint}, arXiv:2310.16028.

\bibitem[{Zhou et~al.(2024)Zhou, Alon, Chen, Wang, Agarwal, and Zhou}]{zhou2024transformersachievelengthgeneralization}
Yongchao Zhou, Uri Alon, Xinyun Chen, Xuezhi Wang, Rishabh Agarwal, and Denny Zhou. 2024.
\newblock \href {https://arxiv.org/abs/2402.09371} {Transformers can achieve length generalization but not robustly}.
\newblock \emph{Preprint}, arXiv:2402.09371.

\bibitem[{Zhu et~al.(2025)Zhu, Dai, and Sui}]{zhu-etal-2025-language}
Fangwei Zhu, Damai Dai, and Zhifang Sui. 2025.
\newblock \href {https://aclanthology.org/2025.coling-main.47/} {Language models encode the value of numbers linearly}.
\newblock In \emph{Proceedings of the 31st International Conference on Computational Linguistics}, pages 693--709, Abu Dhabi, UAE. Association for Computational Linguistics.

\bibitem[{Zhuang et~al.(2023)Zhuang, Liu, Ning, Huang, Lv, Huang, Zhao, Zhang, Mao, Wang et~al.}]{zhuang2023efficiently}
Yan Zhuang, Qi~Liu, Yuting Ning, Weizhe Huang, Rui Lv, Zhenya Huang, Guanhao Zhao, Zheng Zhang, Qingyang Mao, Shijin Wang, et~al. 2023.
\newblock Efficiently measuring the cognitive ability of llms: An adaptive testing perspective.
\newblock \emph{arXiv preprint arXiv:2306.10512}.

\end{thebibliography}
\bibliographystyle{acl_natbib}

\appendix
\section{Example Addition According to Formalization}
\label{appendix:A}
We show a concrete example for two-operand addition according to the formalization defined in Section \ref{subsec:digit10}. For \textbf{$147 + 255$}, we have:

$k=2, d=3, n1 = [1, 4, 7], n2 = [2, 5, 5]$. 

We then compute:

\[T_2 = c_2 + 1 + 2\] \[T_1 = c_1 + 4 + 5\] \[T_0 = c_0 + 7 + 5= 0 + 7 + 5 = 12\] \[s_0 = 12 \mod 10 = 2, \quad c_1 = \left\lfloor \frac{12}{10} \right\rfloor = 1\] \[T_1 = 1 + 4 + 5 = 10\]  \[s_1 = 10 \mod 10 = 0, \quad c_2 = \left\lfloor \frac{10}{10} \right\rfloor = 1\] \[T_2 = 1 + 1 + 2 = 4\]  \[s_2 = 4 \mod 10 = 4, \quad c_3 = \left\lfloor \frac{4}{10} \right\rfloor = 0\] \[S = [0, 4, 0, 2]\]

The result of the addition is $402$. 
\section{Generation Accuracies for 2-Operand, 3-Digit Addition}
\label{appendix:B}
We show the generation accuracy of the full result $S$ and the digit-wise accuracy of $s_2$, compared across the different carry bit datasets, as referenced in Section \ref{subsec:digit10}.
Table \ref{tab:accuracy_2-op} shows that Gemma and Mistral struggle with the generation of the correct result digit $s_2$, exactly in the datasets that \textbf{H1} predicts to be difficult. DS3 and DS5 contain addition tasks in which a lookahead of one digit is insufficient ot determine the value of $s_2$.

\begin{table}[h]
\centering
\small
\begin{tabular}{l|c|c|c|c|c|c}
\toprule
\textbf{}                & \textbf{} & \textbf{DS1} & \textbf{DS2} & \textbf{DS3} & \textbf{DS4} & \textbf{DS5} \\ 
\midrule
&$c_2^h=...$&\(0\)&\(0\)&\(?\)&\(1\)&\(?\)\\
\midrule
\multirow{3}{*}{$\mathbf{S}$} & \textbf{Mistral}      & 0.99               & 1.00               & 0.77               & 1.00               & 0.71               \\ 
& \textbf{Gemma}        &  1.00              & 0.99               & 0.80               & 0.98               & 0.86               \\ 
& \textbf{Llama-3}        & 0.99               & 1.00               & 1.00               & 1.00               & 1.00               \\\midrule
\multirow{3}{*}{\textbf{$\mathbf{s_2}$}} & \textbf{Mistral}       & 1.00               & 1.00               & \textbf{0.77}               & 1.00               & \textbf{0.71}               \\ 
& \textbf{Gemma}         & 1.00               &  0.99              & \textbf{0.81}               & 0.99               & \textbf{0.86}               \\ 
& \textbf{Llama-3}         & 0.99               & 1.00               & 1.00               & 1.00               & 1.00               \\ \bottomrule

\end{tabular}
\caption{Generation accuracy of the full result $S$ and the digit-wise accuracy of $s_2$, compared across the different carry bit datasets.}
\label{tab:accuracy_2-op}
\end{table}
\section{Example: H1 Failure on 4-Operand Addition}
\label{appendix:C}
\begin{figure}[ht]
    \centering
    \includegraphics[width=0.5\textwidth]{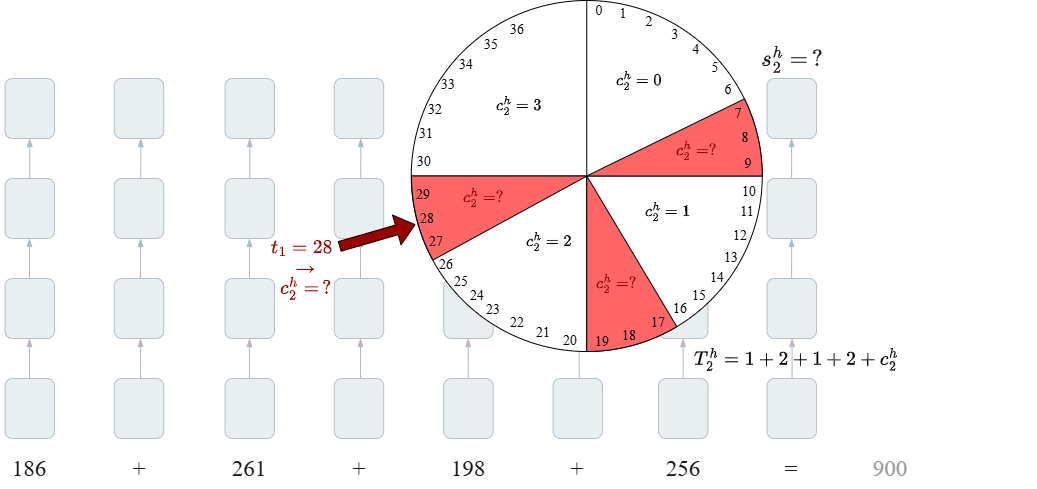} 
    \caption{4-operand addition in which \textbf{H1} fails.} 
    \label{fig:carry_4_op_fail}
\end{figure}

Below is an example in which the heuristic \textbf{H1} fails in 4-operand addition, visualized in Figure \ref{fig:carry_4_op_fail}: 

\noindent\textbf{186 + 261 + 198 + 256.}
\[
\begin{split}
    t_{1} =8 + 6 + 9 + 5 = 28\\
    c_{2}^{h} \in \{ \left\lfloor \frac{c_{min} + 28}{10} \right\rfloor,\\
    \left\lfloor \frac{c_{max} + 28}{10} \right\rfloor \}
\end{split}
\]

with \(c_{max} = 3\)
\[c_{2}^{h} \in \{ \left\lfloor \frac{28}{10} \right\rfloor, \left\lfloor \frac{31}{10} \right\rfloor \} = \{2, 3\}\]
therefore $c_{2}^{h}$ is chosen uniformly at random between $2$ and $3$.
The heuristic thus fails in solving \textbf{186 + 261 + 198 + 256} with a chance of 50\%. 

\section{Zero-shot Generation Accuracy}
\label{sec:appendix_E}

\begin{figure}[h]
    \centering
    \includegraphics[width=0.45\textwidth]{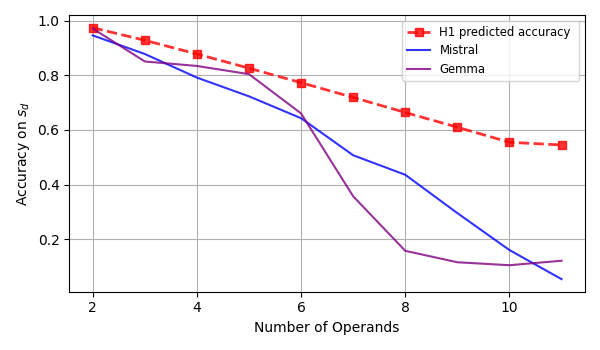} 
    \caption{Accuracy of first generated result digit $s_d$ in zero-shot multi-operand addition tasks for Mistral and Gemma, compared to the expected accuracy on $s_d$ based on \textbf{H1}.} 
    \label{fig:s2_zero-shot}
\end{figure}

We test if \textbf{H1} holds up in predicting the generation accuracy on $s_d$ of Mistral and Gemma for multiple operands. Figure \ref{fig:s2_zero-shot} shows that \textbf{H1} provides an upper bound for the generation accuracy of $s_d$ in a zero-shot setting for Mistral and Gemma on $s_d$.

\section{Accuracy Prediction of Heuristic}
\label{appendix:D}

\renewcommand{\arraystretch}{1.2}
\setlength{\tabcolsep}{5pt}
\begin{table*}[ht]
\centering
\begin{tabular}{|c|c|c|c|c|}
\hline
Nr. Operands $k$ & \textbf{\(c_{max}(k)\)} & Values of \(t_i\) in which H1 fails & Expected acc. on \(s_d\)\\ \hline
2 & 1 & 1 fail:= 9 & $\frac{18\times1.0 + 1 \times 0.5}{19} = 0.974$ \\ \hline
3 & 2 & 4 fails:= 8, 9, 18, 19 & $\frac{24\times1.0 + 4 \times 0.5}{28} = 0.928$\\ \hline
4 & 3 & 9 fails:= 7, 8, 9, 17, 18, 19, 27, 28, 29 & $\frac{28\times1.0 + 9 \times 0.5}{37} = 0.878$ \\ \hline
5 & 4 & 16 fails:= 6, 7, 8, 9, 16, ..., 39 & $\frac{ 30 \times1.0 + 16 \times 0.5}{46} = 0.826$\\ \hline
6 & 5 & 25 fails:= 5, 6, 7, 8, 9, 15, ..., 49 & $\frac{ 30 \times1.0 + 25 \times 0.5}{55} = 0.773$\\ \hline
7 & 6 & 36 fails:= 4, 5, 6, ..., 59 & $\frac{ 28 \times1.0 + 36 \times 0.5}{64} = 0.719$\\ \hline
8 & 7 & 49 fails:= 3, 4, 5, ..., 69 & $\frac{ 24 \times1.0 + 49 \times 0.5}{73} = 0.664$\\ \hline
9 & 8 & 64 fails:= 2, 3, 4, ..., 79 & $\frac{ 18 \times1.0 + 64 \times 0.5}{82} = 0.610$\\ \hline
10 & 9 & 81 fails:= 1, 2, 3, ..., 89 & $\frac{ 10 \times1.0 + 81 \times 0.5}{91} = 0.555$\\ \hline
11 & 9 & 89 fails:= 1, 2, 3, ..., 99 & $\frac{ 10 \times1.0 + 90 \times 0.5}{100} = 0.540$\\ \hline
\end{tabular}
\caption{Predicted accuracy on the first result digit $s_d$ in the addition of multiple numbers according to \textbf{H1}.}
\label{tab:heuristic}
\end{table*}

Table \ref{tab:heuristic} contains, for addition tasks with different numbers of operands $k$, the maximum value of the carry \(c_{max}(k)\). Based on \(c_max\) it list those values of \(t_i\) in which \textbf{H1} is insufficient to accurately predict \(s_2\). Based on the proportion of values of \(t_i\) for which \textbf{H1} is sufficient to the total number of possible values, it lists the predicted accuracy for \(s_2\).

\newpage
\section{Generation Accuracy on All Datasets}
\label{sec:appendix_F}
See Table \ref{tab:gen_accuracy_all}.

\begin{sidewaystable*}[!ht]
\centering
\small
\renewcommand{\arraystretch}{1.0}
\setlength{\tabcolsep}{10pt}
\begin{tabular}{|c|c|c|c|c|c|c|c|c|c|c|c|c|}
\hline
Operands & \multicolumn{4}{c|}{Mistral} & \multicolumn{4}{c|}{Gemma} & \multicolumn{4}{c|}{Llama} \\ \hline
Setting & Overall & $s_2$ & $s_1$ & $s_0$ & Overall & $s_2$ & $s_1$ & $s_0$ & Overall & $s_2$ & $s_1$ & $s_0$ \\ \hline
\multicolumn{13}{|l|}{\textbf{Zero-shot}} \\ \hline
2 & 0.934 & 0.946 & 0.942 & 0.954 & 0.965 & 0.971 & 0.974 & 0.980 & 0.992 & 0.993 & 0.999 & 0.989 \\ \hline
3 & 0.649 & 0.878 & 0.776 & 0.789 & 0.664 & 0.851 & 0.766 & 0.753 & 0.955 & 0.990 & 0.962 & 0.992 \\ \hline
4 & 0.417 & 0.792 & 0.596 & 0.640 & 0.376 & 0.834 & 0.602 & 0.583 & 0.595 & 0.917 & 0.635 & 0.959 \\ \hline
5 & 0.204 & 0.723 & 0.406 & 0.494 & 0.151 & 0.804 & 0.437 & 0.399 & 0.259 & 0.824 & 0.316 & 0.878 \\ \hline
6 & 0.055 & 0.643 & 0.219 & 0.329 & 0.021 & 0.661 & 0.207 & 0.213 & 0.110 & 0.749 & 0.163 & 0.795 \\ \hline
7 & 0.007 & 0.507 & 0.101 & 0.228 & 0.002 & 0.357 & 0.100 & 0.110 & 0.058 & 0.712 & 0.120 & 0.691 \\ \hline
8 & 0.002 & 0.436 & 0.071 & 0.157 & 0.000 & 0.158 & 0.105 & 0.101 & 0.037 & 0.660 & 0.103 & 0.565 \\ \hline
9 & 0.001 & 0.296 & 0.088 & 0.137 & 0.000 & 0.116 & 0.103 & 0.103 & 0.023 & 0.606 & 0.104 & 0.426 \\ \hline
10 & 0.000 & 0.161 & 0.108 & 0.112 & 0.000 & 0.105 & 0.100 & 0.098 & 0.017 & 0.586 & 0.101 & 0.334  \\ \hline
11 & 0.000 & 0.054 & 0.105 & 0.106 & 0.000 & 0.122 & 0.097 & 0.099 & 0.009  & 0.570 & 0.113 & 0.265 \\ \hline
\multicolumn{13}{|l|}{\textbf{One-shot}} \\ \hline
2 & 0.965 & 0.975 & 0.969 & 0.987 & 0.980 & 0.981 & 0.984 & 0.992 & 0.999 & 0.999 & 1.000 & 1.000 \\ \hline
3 & 0.739 & 0.938 & 0.840 & 0.808 & 0.790 & 0.965 & 0.875 & 0.842 & 0.978 & 0.993 & 0.984 & 0.998 \\ \hline
4 & 0.437 & 0.814 & 0.615 & 0.657 & 0.545 & 0.907 & 0.700 & 0.728 & 0.599 & 0.893 & 0.637 & 0.981 \\ \hline
5 & 0.155 & 0.708 & 0.313 & 0.506 & 0.245 & 0.839 & 0.483 & 0.544 & 0.251 & 0.778 & 0.287 & 0.950 \\ \hline
6 & 0.053 & 0.689 & 0.207 & 0.348 & 0.048 & 0.779 & 0.244 & 0.313 & 0.106 & 0.664 & 0.151 & 0.859 \\ \hline
7 & 0.011 & 0.585 & 0.109 & 0.219 & 0.010 & 0.750 & 0.151 & 0.153 & 0.051 & 0.594 & 0.107 & 0.738 \\ \hline
8 & 0.004 & 0.523 & 0.085 & 0.137 & 0.002 & 0.639 & 0.106 & 0.107 & 0.033 & 0.500 & 0.103 & 0.581 \\ \hline
9 & 0.001 & 0.488 & 0.086 & 0.103 & 0.000 & 0.493 & 0.102 & 0.105 & 0.023 & 0.478 & 0.106 & 0.427 \\ \hline
10 & 0.001 & 0.401 & 0.086 & 0.103 & 0.000 & 0.324 & 0.099 & 0.095 & 0.011 & 0.451 & 0.099 & 0.263 \\ \hline
11 & 0.001 & 0.294 & 0.102 & 0.106 & 0.000 & 0.288 & 0.100 & 0.100 & 0.005 & 0.420 & 0.101 & 0.179 \\ \hline
\end{tabular}
\caption{Zero-shot and One-shot settings: Per-digit and overall generation accuracy for all multi-operand addition datasets and models described in Section \ref{sec:models_data}.}
\label{tab:gen_accuracy_all}
\end{sidewaystable*}

\section{Probing Accuracy on Carry Scenarios}
\label{sec:appendix_G}

We evaluate probing accuracy of the probes trained in Section \ref{sec:probing} across the five distinct carry scenarios, introduced in Section \ref{sec:h1_2op}. 

\begin{figure*}[t]
    \centering
    \begin{subfigure}[b]{0.31\textwidth}
        \centering
        \includegraphics[width=\textwidth]{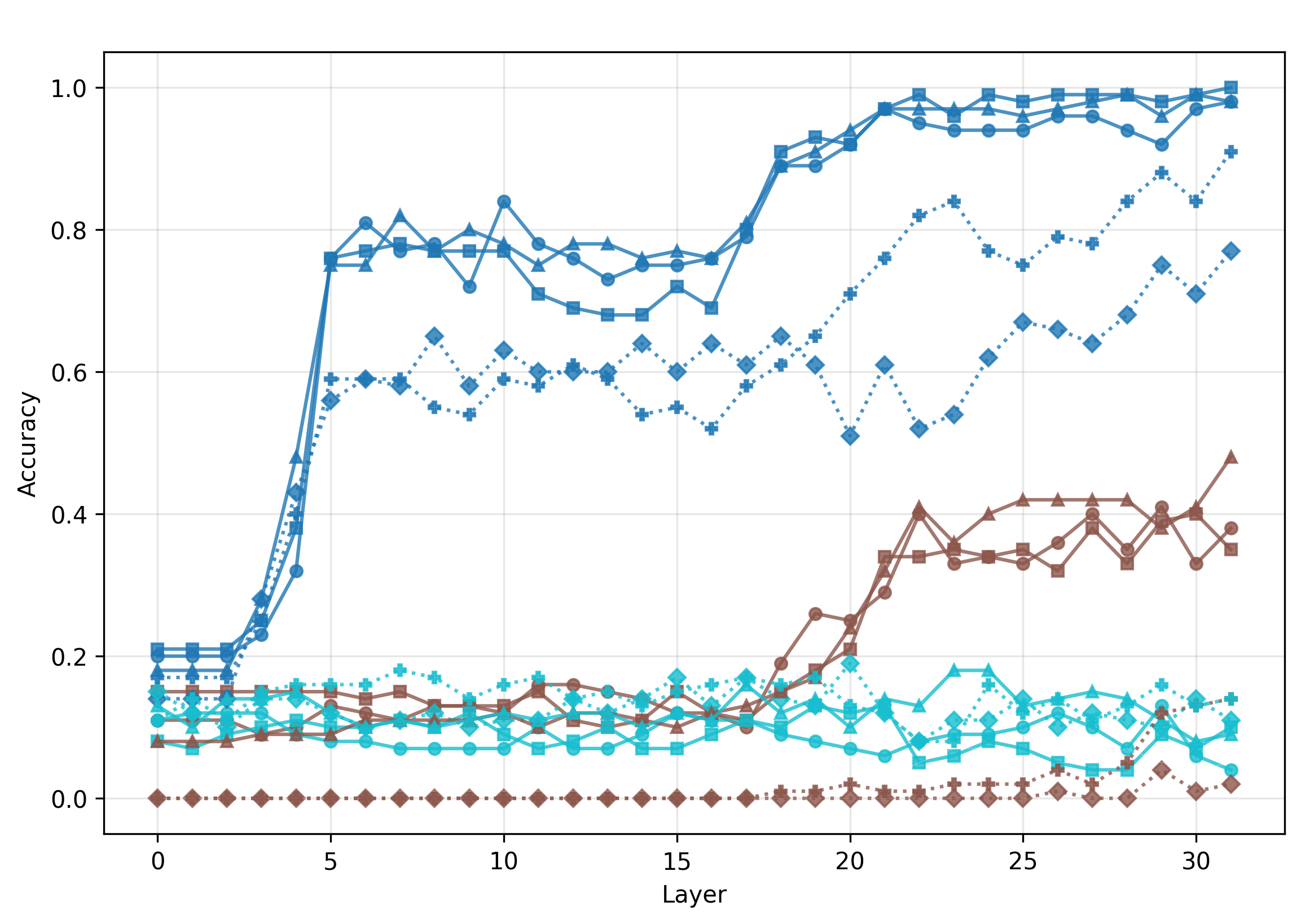}
        \caption{Mistral}
        \label{fig:plot1}
    \end{subfigure} 
    \hfill
    \begin{subfigure}[b]{0.31\textwidth}
        \centering
        \includegraphics[width=\textwidth]{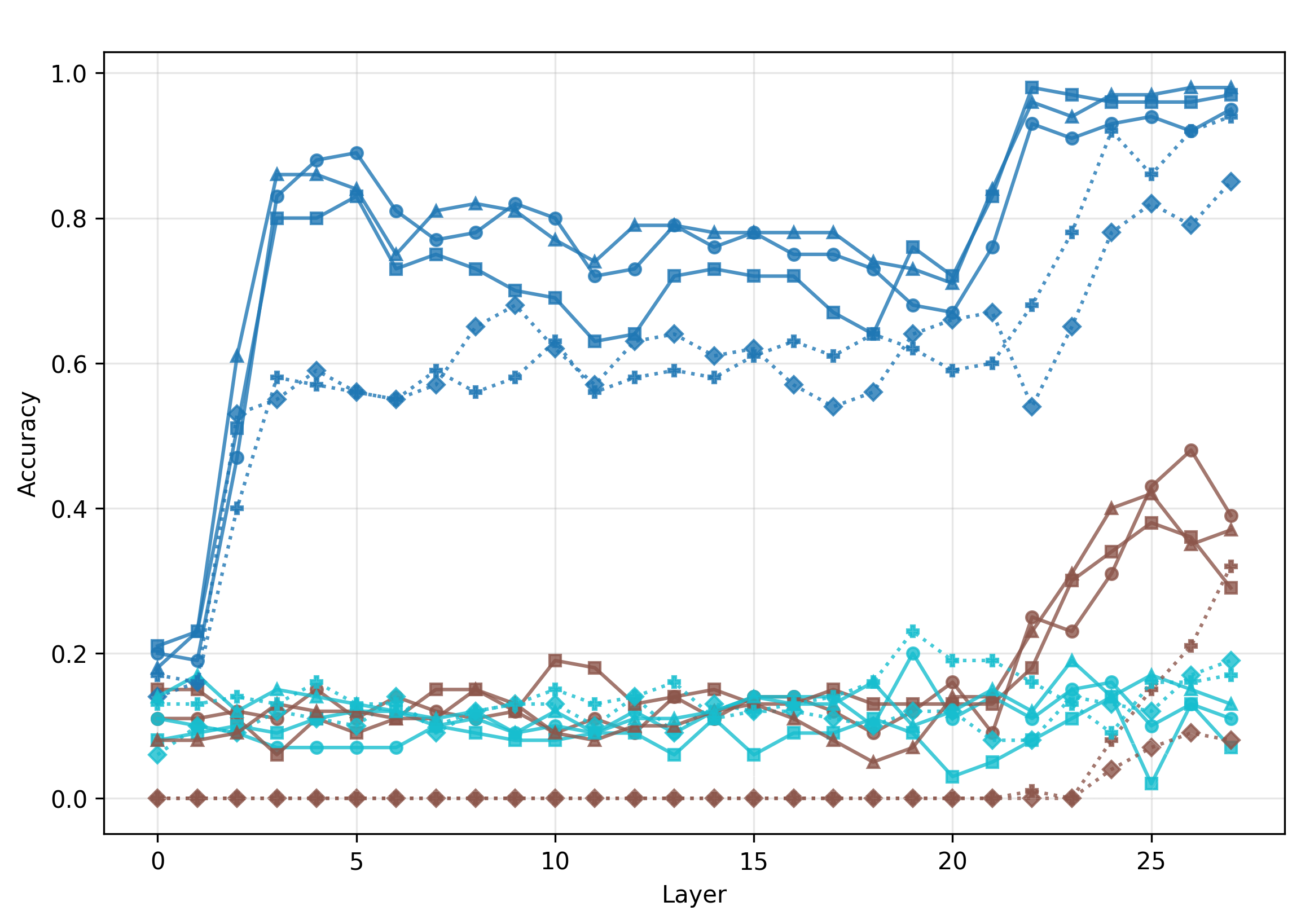}
        \caption{Gemma}
        \label{fig:plot2}
    \end{subfigure}
    \hfill
    \begin{subfigure}[b]{0.31\textwidth}
        \centering
        \includegraphics[width=\textwidth]{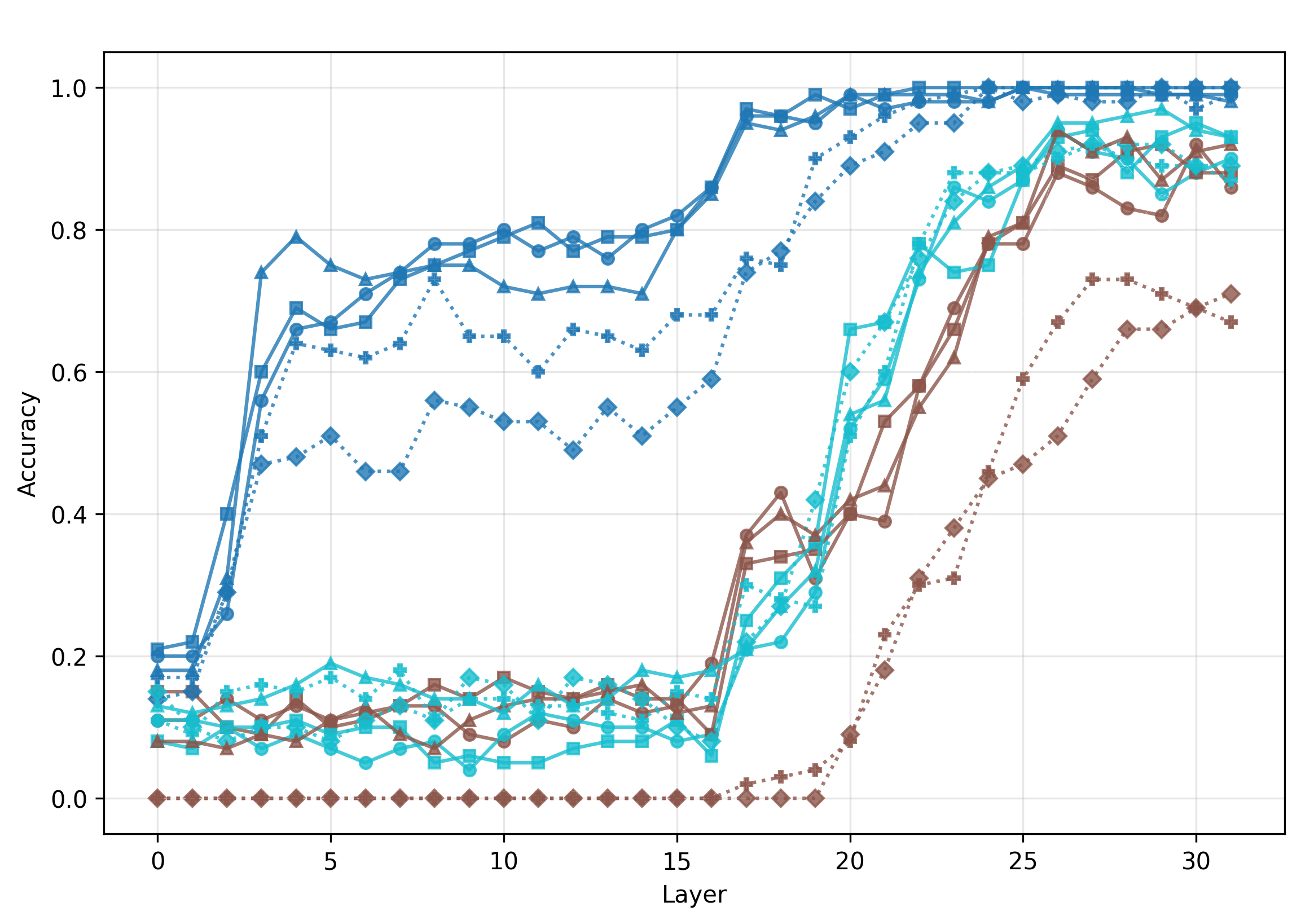} 
        \caption{Llama-3}
        \label{fig:plot3}
    \end{subfigure}
    \hfill
    \begin{subfigure}[b]{0.038\textwidth}
        \centering
        \includegraphics[width=\textwidth]{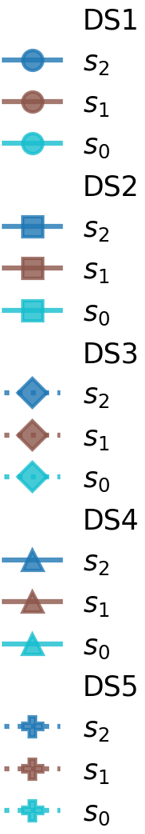} 
    \end{subfigure}
    
    \caption{Digit-wise probing accuracy of result digits of 2-operand addition tasks.
    Each subplot shows the probing accuracies of one model on Datasets DS1-DS5.}
    \label{fig:carry_plot}
\end{figure*}

\paragraph{Results.} Figure \ref{fig:carry_plot} shows that  LLMs struggle with DS3 and DS5, which are exactly the cases where \textbf{H1} would predict problems. The difficult datasets are the ones where a lookahead of one digit position does not suffice to determine the value of the carry needed to generate $s_2$. Simply put: 
In cases where a lookahead of one digit is enough to accurately determine the value of $s_2$ (DS1, DS2, DS4), the models have a relatively good internal representation of the value of the second result digit $s_1$. This results in high performance on the currently generated digit $s_2$. However, when a lookahead of one digit is insufficient to determine the value of $s_2$ (DS3 and DS5), the model struggles with representing digits $s_1$ and $s_2$ correctly.

\end{document}